\documentclass[lettersize,journal]{IEEEtran}
\usepackage{amsmath,amsfonts}
\usepackage{algorithm}
\usepackage{algpseudocode}
\usepackage{array}
\usepackage[caption=false,font=normalsize,labelfont=sf,textfont=sf]{subfig}
\usepackage{textcomp}
\usepackage{enumitem}
\usepackage{stfloats}
\usepackage{url}
\usepackage{verbatim}
\usepackage{graphicx}
\usepackage{cite}
\usepackage{xcolor} 
\usepackage{multirow}
\usepackage{adjustbox}
\DeclareMathAlphabet\mathbfcal{OMS}{cmsy}{b}{n}
\newcommand{\ten}[1]{\mathbfcal{#1}}
\newcommand{\mat}[1]{\mathbf{#1}}
\colorlet{blue}{black}
\usepackage{multicol,lipsum}
\usepackage{svg}
\begin{document}

\title{Ultra Memory-Efficient On-FPGA Training of Transformers via Tensor-Compressed Optimization}

\author{Jiayi Tian, Jinming Lu, Hai Li, Xiangwei Wang, Cong (Callie) Hao, Ian Young, \textit{Fellow, IEEE}, Zheng Zhang
        % <-this % stops a space
\thanks{J. Tian, J. Lu and Z. Zhang are with Department of Electrical and Computer Engineering, University of California, Santa Barbara, CA 93106; H. Li and I. Young are with Intel Corporation, Hillsboro, OR 97124; X. Wang is with Department of Computer Science, North Carolina State University. He participated in this work when he was an exchange undergraduate student at UCSB; C. Hao is with School of Electrical and Computer Engineering, Georgia Institute of Technology, Atlanta, GA 30332. She participated in this work before Intel funded this project. (email: jiayi\_tian@ucsb.edu; jinminglu@ucsb.edu; hai.li@intel.com; xwang258@ncsu.edu; callie.hao@gatech.edu; ian.young@intel.com; zhengzhang@ece.ucsb.edu).}

}

% \markboth{Journal of xxx,~Vol.~xxx, No.~xxx, xxx~xxx}%
% {Shell \MakeLowercase{\textit{et al.}}: Ultra Memory-Efficient On-FPGA Training of Transformers via Tensor-Compressed Optimization}

%\IEEEpubid{0000--0000/00\$00.00~\copyright~2025 IEEE}

\maketitle

\newcommand{\zz}[1]{{\color{red}[zz: #1]}}

\begin{abstract}
% Transformer & training Transformer on-edge applications
Transformer models have achieved state-of-the-art performance across a wide range of machine learning tasks. There is growing interest in training transformers on resource-constrained edge devices due to considerations such as privacy, domain adaptation, and on-device scientific machine learning. 
However, the significant computational and memory demands required for transformer training often exceed the capabilities of an edge device.
% There have been multiple previous works for on-edge inference of Transformer trained in the cloud, 
% However, there has not been an on-device training engine for Transformer models because its large amount of parameters introduces high memory and computation costs. 
% To tackle these difficulties, there have been quite a lot of methods proposed for compressing Transformer models, such as knowledge distillation, quantization, pruning, and tensor decomposition. 
% Nevertheless, most of these methods could only reduce memory and computation burden in the inference stage, and could even increase the training cost. 
% For instance, knowledge distillation introduces teacher and student models and runs forward simultaneously, doubling the memory and operation usage in the training process.
% Among them, tensor decomposition is a method that could reduce the model volume both in training and inference by directly decomposing weight matrices into small high-dimensional tensors. 
Leveraging low-rank tensor compression, this paper presents the first on-FPGA accelerator for transformer training.
%Tensor decomposition could directly reduce the model volume in training by decomposing weight matrices into small high-dimensional tensors and training from scratch. Benefiting from the hardware-efficient tensor contraction operations, some previous works design hardware accelerators for tensorized neural networks, but only for model inference. To better leveraging tensor decomposition for training on edge, this work proposes a fast and memory-saving tensorized Transformer training accelerator, with a co-design of optimized computing flow and tensor contraction kernel.
On the algorithm side, we present a bi-directional contraction flow for tensorized transformer training, significantly reducing the computational FLOPS and intra-layer memory costs compared to existing tensor operations.
On the hardware side, we store all highly compressed model parameters and gradient information on chip, creating an on-chip-memory-only framework for each stage in training. This reduces off-chip communication and minimizes latency and energy costs. Additionally, we implement custom computing kernels for each training stage and employ intra-layer parallelism and pipe-lining to further enhance run-time and memory efficiency.
% to avoid additional memory creation for weight transpose required in backward propagation, we combine the usage of inner product and row-wise product to enable rank-dimension parallelism for accelerating multiple training stages.
Through experiments on transformer models within $36.7$ to $93.5$ MB using FP-32 data formats on the ATIS dataset, our tensorized FPGA accelerator could conduct single-batch end-to-end training on the AMD Alevo U50 FPGA, with a memory budget of less than $6$-MB BRAM and $22.5$-MB URAM. 
Compared to uncompressed training on the NVIDIA RTX 3090 GPU, our on-FPGA training achieves a memory reduction of $30\times$ to $51\times$. Our FPGA accelerator also achieves up to $4.0\times$ less energy cost per epoch compared with tensor transformer training on an NVIDIA RTX 3090 GPU.
% \zz{The memory saving over uncompressed training is xxx}. 
As an initial result, this work highlights the significant potential of large-scale tensor training on edge devices.
% Finally, for the bottom-level tensor contraction kernel design, we use two types of GEMM cores, including inner product and row-wise product to support parallel computing for unified tensorized weight storage in forward and backward propogations.
% A cyclic interleaving memory reshaping approach is proposed to enable parallel computing for unified tensorized weight storage in each training stage.
% Additionally, we devise a unified-rank hardware-friendly training algorithm considering both training accuracy and resource usage.  
% Additionally, to combine the advantages of multi-head attention and tensorized linear layer, we proposed a fused tensor-attention kernel to further increase the memory efficiency.
%Additionally, to support layer extension for the Transformer training engine, we encourage on-chip memory reuse for intermediate results generated by different layers for memory efficiency.
\end{abstract}


\newcommand{\jiayi}[1]{{\color{blue}[jiayi: #1]}}

\section{Introduction}

\IEEEPARstart{T}{here} has been growing interest in on-device training of machine learning models. 
In finance and healthcare \cite{liu2021machine, papernot2018sok}, \textcolor{blue}{privacy regulations and user expectations often prohibit raw data from being transmitted to external servers, making local training a practical and necessary solution.}
% This has driven federated learning~\cite{wei2020federated,li2021survey} on edge devices to train AI models end-to-end or incrementally based on local private data. 
In robotic and autonomous systems, machine learning has shown great promise in high-dimensional safety verification and control~\cite{bansal2021deepreach,onken2020neural,sun2021learning}, where the neural network needs to be trained again as the system dynamics or safety regions keep updated. 
Meanwhile, pre-trained AI model often suffers from performance degradation in practical deployment due to the uncertain environment and data distribution shift~\cite{lee2020learning}, necessitating on-device model adaptation. 
Due to the limited size, weight, and power budget, resource-constrained devices such as FPGA, micro-controllers, or embedded GPUs are often used for edge AI. 
In recent years, several edge training accelerators have been proposed for convolutional neural networks \cite{tang2022ef, guo2023boost}, multilayer perceptron networks, or recurrent neural networks~\cite{lu2022eta, fox2019training, zhang2022fast}. 
Most existing work focuses on on-device \textcolor{blue}{lightweight adaptation or partial fine-tuning}, \textcolor{blue}{rather than the more challenging—but potentially more effective—full-model training}, due to the limited computing and memory budget. 

Transformer models~\cite{vaswani2017attention} have achieved state-of-the-art performance in numerous application domains, including (but not limited to) natural language processing \cite{vaswani2017attention, kenton2019bert}, computer vision \cite{dosovitskiy2020image, kirillov2023segment} and scientific simulation~\cite{zhaopinnsformer}. 
\textcolor{blue}{Recently, on-device fine-tuning of transformers has been explored in practical settings such as multilingual voice assistants and document scanners, where adapting models to local data can enhance personalization while preserving user privacy~\cite{yu2021federated, jia2022federated}. These scenarios underscore the growing demand for efficient transformer training directly on edge devices—particularly in applications where centralized retraining is impractical or undesirable.}
\textcolor{blue}{Compared with on-device fine-tuning, end-to-end on-device training is more important in physics intelligence (e.g., neural PDE solvers for safe robotic control) since the solution will change dramatically as sensor data and system dynamics change, leading to complete failure of fine-tuning.}
However, the substantial memory and computational burdens of transformers have posed new challenges for their training and deployment on resource-constrained devices. 
% However, due to the rising concerns about data privacy, the desire to reduce dependency on centralized computing infrastructures, and the growing demand for models to adapt to dynamic environments and data uncertainty, 
For instance, the model parameters of a medium-size transformer can easily exceed the memory capacity of an edge device (e.g. FPGA), making it nontrivial to perform inference and infeasible to perform  training on it.
%However, Field Programmable Gate Arrays (FPGAs) are often preferred in resource-constrained scenarios, due to their re-configurability, energy efficiency, and low hardware cost. Meanwhile, FPGA prototypes can provide a solid proof-of-concept demos for more powerful (yet more expensive) ASIC AI accelerators.
% Consequently, enabling transformer model training on FPGA is an intriguing research topic.

Recent works have investigated quantization \cite{tian2023bebert,ji2024beta,wang2018training}, low-rank compression \cite{yang2023quantization, yang2024loretta, yang2024adazeta}, pruning \cite{fang2022algorithm,fang2023efficient} and knowledge distillation \cite{wang2020minilm, yangwanda++} to reduce the cost of fine-tuning on GPU or inference on edge devices for transformer models.  
However, full-model transformer training on resource-constrained devices remains a great challenge, due to two main factors:
\begin{itemize}[leftmargin=*]
    \item Firstly, it is hard to meet the memory requirement of transformer training. A transformer model can easily consume $>5\times$ more on-chip memory than the capacity of an FPGA. It is difficult to fill this memory gap even if sparse~\cite{rhu2018compressing} or low-precision training~\cite{hubara2017quantized,gupta2015deep,sun2020ultra,wang2018training} methods are utilized. For example, ideal memory reduction (which is hard to achieve in practice) is limited to $8\times$ even if ultra-low-precision 4-bit training~\cite{sun2020ultra} is used.
    \item Secondly, on-device training involves more computing tasks than on-device inference, such as backward propagation, gradient generation, and parameter update. This increases both the memory and computing burden on the edge devices. Furthermore, data dependency between forward and back propagations further complicates data communication, task scheduling, and memory management.
\end{itemize}

% \begin{figure}[htbp]
%     \centering
%     \includegraphics[width=\linewidth]{fig_new/occupancy_memory.pdf}
%     \caption{\textcolor{blue}{Comparison of GPU usage efficiency of matrix-based and tensor-train-based transformer training. Results are measured on the training task of a two-encoder transformer using torch profile tools.}}
%     \label{fig:gpu-profile}
% \end{figure}

% To address the above challenges, we propose an efficient edge training accelerator for transformer-based models utilizing tensor decomposition. 
% Compared to the quantization approach, tensor decomposition has shown order-of-magnitude compression ratio in multiple tasks. 
% In this paper, we use FPGA as a demonstration to investigate the end-to-end training of Transformer models on edge devices that leverage low-rank tensor optimization. 
In this paper, we use FPGA as a demonstration platform to investigate the end-to-end training of transformer models on edge devices that leverage low-rank tensor optimization. 
As a high-dimensional generalization of matrix decomposition~\cite{kolda2009tensor}, tensor decomposition can achieve much higher compression ratios in neural networks. This may provide a huge reduction in memory and computing costs, enabling on-device training and inference of large AI models that were previously infeasible. 
Previous work has shown orders-of-magnitude model reduction \textcolor{blue}{in convolutional and recurrent networks} by low-rank tensor decomposition \textcolor{blue}{during both post-training compression~\cite{lebedev2015speeding, kim2015compression, zhen2019fast} and training-aware compression~~\cite{novikov2015tensorizing,calvi2019compression,hrinchuk2020tensorized}. }
\textcolor{blue}{For transformer models, some studies have applied tensor decomposition to compress specific components such as embedding or attention layers~\cite{hrinchuk2020tensorized, ma2019tensorized}, but these approaches target only partial modules and achieve limited model compression ratio.}
% Other works such as~\cite{yang2024loretta,yang2024adazeta, ghiasvand2024communication} use tensorized adapters for parameter-efficient fine-tuning of large language models (LLMs). \textcolor{blue}{However, these adapter-based approaches reduce only trainable parameters, and do not reduce the model size—limiting their applicability in memory-constrained edge devices.}

\textcolor{blue}{From a hardware execution standpoint, although tensor decomposition offers theoretical reductions in memory footprint and computation, practical speedup on GPUs is often limited by the tiny and sequential tensor-contracted operations. We profile the top-10 most time-consuming kernels during MM- and TT-based transformer training, where we found that TT kernels consistently exhibit $6.5\times$ lower occupancy and $3.0 \times$ fewer blocks per streaming multiprocessor (SM) on average. This leads to inefficient scheduling, underutilized GPU resources, and increased latency.
In contrast, FPGAs support custom dataflows and explicit memory control, making them more suitable for executing tensor-compressed transformer training efficiently on edge devices.
While prior FPGA/ASIC~\cite{deng2019tie, gong2023ette, meng2023tt, guo202115}, emerging computing platforms \cite{zhao2023tensor, zhao2023tensorized, zhaoreal} designs have explored this potential, most focus solely on inference or small-model training acceleration, which do not address full-model on-device training for transformer models.}

% Some recent work has shown high accuracy of rank-adaptive tensor-compressed training~\cite{yang2024comera,hawkins2022towards, hawkins2021bayesian}, speedup over standard pre-training on Transformers and large language models (LLM) via GPU optimization~\cite{yang2024comera}, as well as memory savings and communication reduction for LLM fine-tuning~\cite{yang2024loretta,yang2024adazeta, ghiasvand2024communication}. 
% Using these algorithmic results, some previous work also developed tensor-compressed methods without backpropagation to train small neural networks on photonic platforms~\cite{zhao2023tensor,zhao2023tensorized}. 
% However, current back-propagation-free training cannot handle Transformer-size models due to its large gradient estimation error, and on-device \textcolor{blue}{full-model} Transformer training remains an open problem. 

\begin{figure}[t]
     \centering
     \includegraphics[width=.45\textwidth]{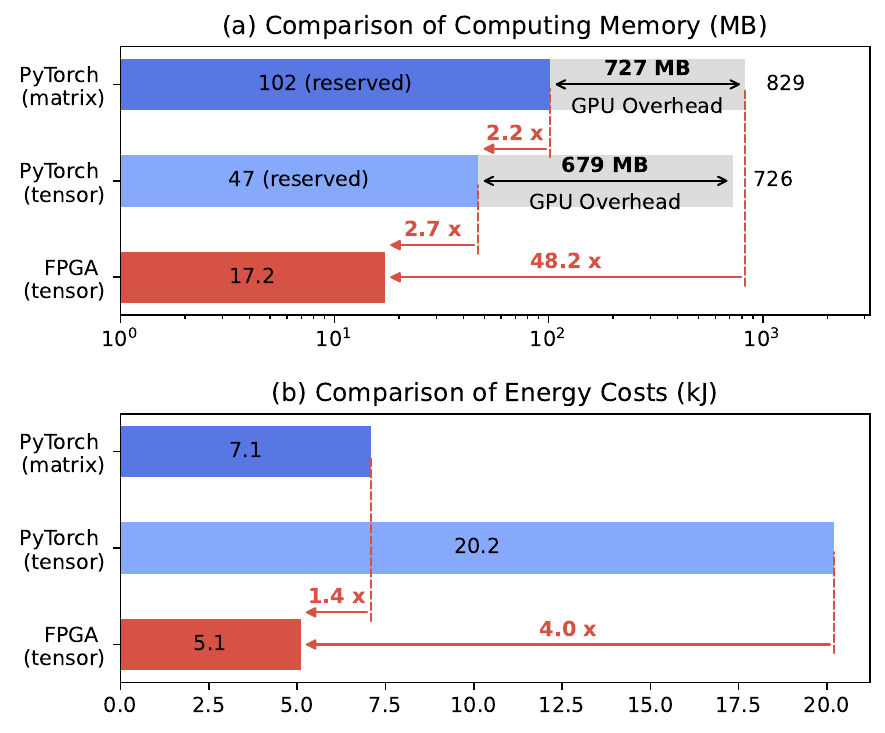}
    \caption{\textcolor{blue}{Comparison of computing memory consumption (top) and energy costs (bottom) between PyTorch-based GPU training and our accelerator; blue bars represent reserved CUDA memory without additional overhead. }} 
    % \zz{You misunderstood my comments. (1) use the original figure on the left for memory comparison (with 5 bars), (2) add another figure to the right (with 3 bars) to show the energy cost. Do NOT put the results of two performance metrics in one figure: they overlap with each other, and it's hard to distinguish them.}\jiayi{I am trying to make the figure more clear by using three hists since the memory / energy could have a better correspondence.}}
    % \zz{add another sub-figure to the right showing the energy per epoch: matrix-based on GPU, tensor-based on GPU, tensor-based on FPGA. }\jiayi{i am not sure how to draw the other figure since one has 5 hists and the other has 3 hists, which is unbalanced, and should the combined figure take the whole page or half page?} \zz{It's fine that the other paper only has three bars. As long as the two figures have the same height and width, it would be fine.}}
    % Make sure that you use the new model setting adopted in in your improved FPGA design.} \
    \label{fig:intro}
\end{figure}

To fill the research gap described above, this paper presents, for the first time, an FPGA accelerator for low-rank tensor-compressed {\it full-model} transformer training. This research can be regarded as a proof-of-concept demonstration and a pathfinding effort toward future ASIC implementation. This article presents the following novel contributions:
\begin{itemize}[leftmargin=*]
    \item We design a tensor-compressed transformer training accelerator implemented on FPGA. Using tensor decomposition, transformer training becomes ultra memory-efficient, allowing full-model training of up to $93.5$-MB model with an $28.4$-MB on-chip memory budget.
    % [including less than $6$-MB block RAM (BRAM) capacity]. 
    % Without using any quantization or pruning techniques, 
    % This ultra memory-efficient method can finish end-to-end transformer training on an FPGA with less than 28.4-MB on-chip memory budget. 
    % which significantly alleviate storage and computational burdens without sacrificing the accuracy for transformer training on resource-constrained devices. 
    \item We present a novel bidirectional tensor contraction technique that significantly improves the computing and memory efficiency of tensor-compressed forward and backward propagation. This contraction flow surpasses the sequential computation flows employed in previous inference accelerators~\cite{deng2019tie,gong2023ette}. In addition, we develop a comprehensive model for analyzing the computing and memory costs associated with various tensor decomposition formats and contraction orders.
    \item We optimize the parallel scheduling flow and develop an operation fusion strategy to enhance the trade-off between run-time and hardware cost. Additionally, we propose a tensor grouping technique to enhance the memory efficiency of storing tensor-compressed model weights.
    \item We implement the training algorithm on an AMD Alveo U50 FPGA, evaluate hardware performance using the ATIS dataset~\cite{hemphill1990atis} on a transformer model with various number of encoding blocks. Experimental results demonstrate that our FPGA accelerator can achieve up to $4.0\times$ lower energy costs with $48.2\times$ reduction in computing memory compared to transformer training using Pytorch on a Nvidia RTX 3090 GPU, as highlighted in Fig.~\ref{fig:intro}.
    
    % \zz{Need to show the following: memory reduction from full-size training on GPU to tensorized training on GPU, then from tensorized training on GPU to tensorized training on FPGA. Better to have a figure to show the step-by-step reduction at the end of the introduction section.}
    % \item We propose an on-chip-only framework for each computing phase in training, which eliminates the additional off-chip communication to lower latency and energy costs.
    % \item We enable parallelism and pipeline among multiple contraction steps inside tensor linear layer combining with our bi-directional computing flow to further reduce the time complexity and memory burden.
\end{itemize}
As a preliminary study, this work has not used any pruning or quantization techniques, and there hardware resource is still under utilized. Therefore, the performance gain can be further improved in the future. However, the current initial study suffices to demonstrate the great potential of low-rank tensor computation to scale up training on edge devices.

\begin{figure*}[t]
\includegraphics[width=\textwidth]{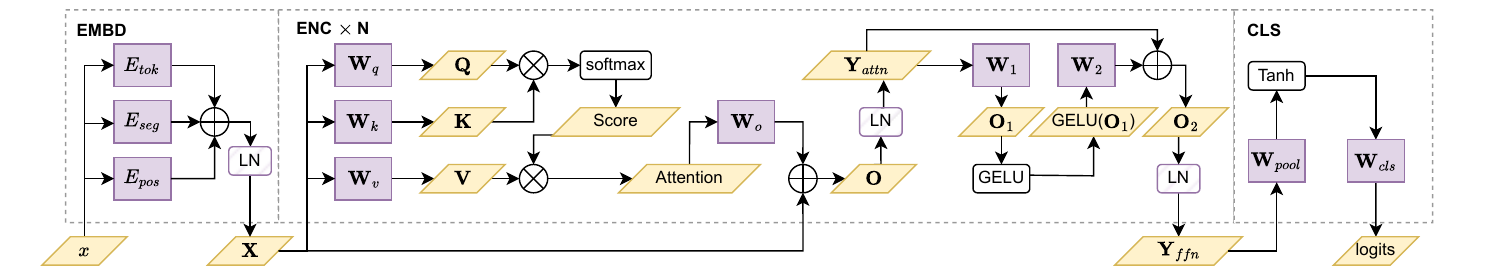}
\caption{Transformer structure for classification tasks. Inter-layer activation is represented using yellow blocks, embedding tables and linear layer weights are in purple blocks, and non-linear functions are in white blocks.}
\label{fig:Trans}
\end{figure*}

\section{Background}
In this section, we introduce some necessary background related to our work, \textcolor{blue}{including the transformer architecture, tensor decomposition and contraction, and tensor-compressed neural network.}

\subsection{Transformer Models}
% \zz{Please add a short introduction/tutorial about transformers (e.g., for half a page), including one figure to show the architecture. Around 1/3 page.}

% \zz{Please add a figure to show transformer architecture, and use a few equation to explain the key operations of each block (e.g., embedding, attention, feed-forward network). This should be written like a short lecture rather than literature review.}

% Transformers were initially introduced in NLP to address limitations in traditional sequential models like recurrent neural networks (RNNs) by allowing for the parallel processing of data. They rely on an attention mechanism that selectively focuses on different parts of the input, which enables better modeling of long-range dependencies. This is particularly useful for tasks such as machine translation \cite{vaswani2017attention, kenton2019bert} and text summarization \cite{zhang2019pretraining, liu2019text}. Unlike earlier CNN models, transformers do not rely on fixed-size windows or filters, which makes them adaptable to a variety of input sizes and structures. This flexibility has enabled transformers to outperform other architectures in various tasks and has led to their adoption in fields beyond NLP, including computer vision, where they are used for tasks such as object detection \cite{beal2020toward, carion2020end} and image classification \cite{dosovitskiy2020image, chen2021crossvit}. 

A transformer model consists of a stack of encoder or decoder blocks. 
For discrimination tasks, one can use encoders only to process the input sequence and generate output representations. 
For generative tasks, decoders should be used to generate the output sequence step by step, using both the encoded input and its own previous output. 
A encoder block and a decoder block have the same structure, a self-attention layer, and a feed-forward network (FFN). 
Both encoder and decoder blocks include residual connections and layer normalization to stabilize training. 
The computation in one encoder block could be  written sequentially as
% \begin{align}
% &\mat{Q} = \mat{W}_q \mat{X} + \mat{B}_q, \mat{K} = \mat{W}_k\mat{X} + \mat{B}_k, \mat{V} = \mat{W}_v\mat{X} + \mat{B}_v, \\
% &{\rm Attention}(\mat{Q,K,V}) = {\rm softmax}(\frac{\mat{Q}\mat{K}^T}{\sqrt{d_k}})\mat{V},  \\
% &\mat{Y}_{\rm attn} = {\rm LN}(\mat{W}_{\rm o}\ {\rm Attention}(\mat{Q,K,V}) + \mat{B}_{\rm o} + \mat{X}),  \\
% &\mat{Y}_{\rm ffn} = {\rm LN}\left (\mat{W}_2\ {\rm GELU}\left( \mat{W}_1 \mat{Y}_{\rm attn}+\mat{B}_1 \right)+\mat{B}_2+ \mat{Y}_{\rm attn} \right).  
% \end{align}
\begin{align}
\mathbf{Q} = \mathbf{W}_q \mathbf{X} + \mathbf{B}_q, 
\mathbf{K} = \mathbf{W}_k \mathbf{X} + \mathbf{B}_k, 
\mathbf{V} = \mathbf{W}_v \mathbf{X} + \mathbf{B}_v, \notag 
\end{align}
\begin{align}
\text{Attention}(\mathbf{Q}, \mathbf{K}, \mathbf{V}) = \text{Softmax}\left( \frac{\mathbf{Q} \mathbf{K}^\top}{\sqrt{d_k}} \right) \mathbf{V}, 
\end{align}
\begin{align}
\mathbf{Y}_{\text{attn}} = \text{LN}\left( \mathbf{W}_o \, \text{Attention}(\mathbf{Q}, \mathbf{K}, \mathbf{V}) + \mathbf{B}_o + \mathbf{X} \right), \notag
\end{align}
\begin{align}
\mathbf{Y}_{\text{ffn }} = \text{LN}&\left( \mathbf{W}_2 \, \text{GELU}(\mathbf{W}_1 \mathbf{Y}_{\text{attn}} + \mathbf{B}_1) + \mathbf{B}_2 + \mathbf{Y}_{\text{attn}} \right). \notag
\end{align}
\textcolor{blue}{Let $d_{\text{hid}}$ denote the hidden state dimension of the encoder, and let $N_{\text{seq}}$ be the sequence length.}
Here, $\mat{X}$ is the input to the encoder, and $\mat{Y}_{\text{attn}}$ represents the output of the attention layer and feed-forward network, both with shape \textcolor{blue}{$d_{\text{hid}} \times N_{\text{seq}}$}.
The matrices $\mat{W}_q$, $\mat{W}_k$, $\mat{W}v$, and $\mat{W}o$ are the projection weights in the attention mechanism, while $\mat{W}1$ and $\mat{W}2$ are the weights in the feed-forward network, each with shape \textcolor{blue}{$d_{\text{hid}} \times d_{\text{hid}}$}.
The matrices $\mat{Q}$, $\mat{K}$, and $\mat{V}$ denote the query, key, and value features, respectively, each of shape \textcolor{blue}{$d_{\text{hid}} \times N_{\text{seq}}$}.
${\rm LN}$ denotes the layer normalization operation. The nonlinearities used are ${\rm Softmax}$ in the attention mechanism and ${\rm GELU}$ in the feed-forward network.

\textcolor{blue}{Before the encoder}, an embedding layer transforms discrete tokens (e.g., words or subwords) into dense vectors for semantic processing by the attention mechanism.
To encode positional information, positional embeddings ${\rm E}_{\rm pos} \in \textcolor{blue}{\mathbb{R}^{d_{\text{hid}} \times N_{\text{seq}}}}$ are added to the token representations.
In natural language tasks, token embeddings ${\rm E}_{\rm tok} \in \textcolor{blue}{\mathbb{R}^{d{_\text{hid}} \times |\mathcal{V}|}}$ map tokens $x_i$ from a vocabulary $\mathcal{V}$ into continuous vectors, while segment embeddings ${\rm E}_{\rm seg} \in \textcolor{blue}{\mathbb{R}^{d_{\text{hid}} \times S}}$ (where $S$ is the number of segments, e.g., sentence A and B) distinguish tokens from different segments, enabling sentence-pair modeling and similar tasks.
% Secondly, segment and token embeddings are required in natural language processing to process an input sequence.
% The token embedding ${\rm E}_{\rm tok} \in \textcolor{blue}{\mathbb{R}^{d_{\text{hid}} \times |\mathcal{V}| }}$ map input tokens $x_i$ from a vocabulary $\mathcal{V}$ into continuous vector representations.
% Segment embeddings ${\rm E}_{\rm seg} \textcolor{blue}{ \in \mathbb{R}^{d_{\text{hid}} \times S}}$ (where $S$ is the number of segments, e.g., sentence A and B) distinguish tokens from different segments, which is essential for tasks like sentence-pair classification.
The embedding output $\textcolor{blue}{z_i \in \mathbb{R}^{d_{\text{hid}} \times N_{\text{seq}}}}$ of a transformer can be expressed as
\begin{align}
    z_i = {\rm E}_{\rm tok}(x_i) + {\rm E}_{\rm pos}(i) + {\rm E}_{\rm seg}(p_i), 
\end{align}
where $x_i$ is the $i$-th token in the input sequence \textcolor{blue}{and $p_i$ indicates its segment ID}.

For classification tasks, a classifier—typically consisting of one or more linear layers followed by a non-linear activation—is applied to the final hidden representation of the [CLS] token to perform prediction.
Fig.~\ref{fig:Trans} illustrates the model structure of a standard transformer used for classification.

\subsection{Tensor Decomposition and Contraction}

% \zz{Do Not talk about tensorized neural network training here. Introduce what tensors, tensor decomposition (especially tensor train and TTM), and tensor networks (using figures) are. See the background section of the CoMERA paper. See also the tensor network diagram in my slides used for Intel annual project review. Around 1/2 page. }

Tensors are a high-dimensional generalization of matrices and vectors. A generic $d$-way could be represented as $\ten{A}\in \mathbb{R}^{n_1\times ...\times n_d}$, where $n_k$ is the size of the mode (or dimension) $k$. The $(i_1, i_2, \cdots, i_d)$-th element of $\ten{A}$ is written as $a_{i_1, i_2, \cdots, i_d}$ or $\ten{A}(i_1, i_2, \cdots, i_d)$. One can also use graphs to represent tensor operators. Specifically, a $d$-way tensor is denoted by a node with $d$ edges. If two nodes are connected by an edge, then a tensor contraction happens between them, and the connected mode will diminish. Consider, for example, $\ten{A}\in \mathbb{R}^{n_1\times ...\times n_d}$ and $\ten{B}\in \mathbb{R}^{m_1\times ...\times m_l}$. Assume that $n_s = m_t$, the tensor contraction of these two tensors could be formulated as 
\begin{align}
    \ten{C} = & \ten{A}\times_s^t\ten{B}. 
\end{align}
Tensor $\ten{C}\in \mathbb{R}^{n_1\times \cdots n_{s-1} \times n_{s+1} \cdots \times n_d \times m_1 \times \cdots m_{t-1} \times m_{t+1} \cdots \times m_l}$ has $d+l-2$ modes, and each element of $\ten{C}$ is
\begin{align}
      & c_{i_1, \cdots, i_d, j_1, \cdots, j_l} = \sum \limits_{i_s, j_t=1}^{n_s} a_{i_1, \cdots, i_{s}, \cdots, i_d} b_{j_1, \cdots, j_t, \cdots, j_l}
\end{align}
The contraction operations of multiple tensors can be represented as a tensor network with many nodes and edges. The number of free edges in a tensor network decides the final shape and size of the resulting tensor. 

It is expensive to store and compute high-dimensional tensors directly. Fortunately, practical tensors can often be compressed by low-rank tensor decompositions~\cite{kolda2009tensor}. We mainly use the tensor-train (TT) decomposition~\cite{oseledets2011tensor} and its variant tensor-train-matrix (TTM) decomposition for compressed training.
TT decomposes a tensor $\ten{A}\in \mathbb{R}^{n_1 \times ... \times n_d}$ as 
\begin{align}
\ten{A} = \ten{G}_1\times_3^1\ten{G}_2\times_3^1...\ten{G}_d,
\end{align}
where $\ten{G}_k\in \mathbb{R}^{r_{k-1}\times n_k\times r_{k}}$ is a TT factor, $(r_0, r_1, \cdots, r_d)$ is called TT rank, and $r_0 = r_d = 1$. 
TTM decomposes an order-$2d$ tensor $\ten{W}\in \mathbb{R}^{n_1 \times ... \times n_d \times m_1\times ... \times m_d}$ as
\begin{align}
\label{eq: ttm}
\ten{W} = \ten{F}_1\times_4^1\ten{F}_2\times_4^1...\ten{F}_{d},
\end{align}
where $\ten{F}_k\in \mathbb{R}^{r_{k-1}\times n_{k} \times m_{k} \times r_{k}}$, and $r_0 = r_d = 1$.  

\begin{figure}[t]
\centering
\includegraphics[width=.45\textwidth]{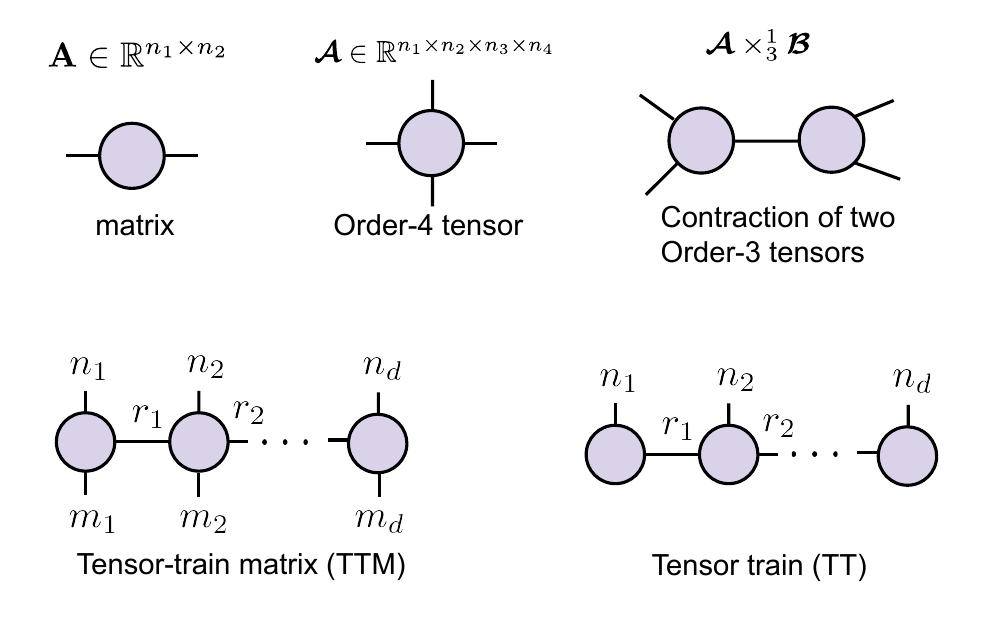}
\caption{Tensor graph representation for tensors, tensor contraction, and tensor networks for TTM and TT decomposition.}
\label{fig:tg_basic}
\end{figure}

The graph representations for tensors, tensor contractions, and tensor networks for TTM and TT are shown in Fig. \ref{fig:tg_basic}.

\subsection{Tensor-Compressed Neural Network}
\label{sec:low-rank}
% \zz{Do NOT talk about transformer training. Talk about in general how tensor decomposition can be used in DNN training: (1) TT for FC layers, (2) TTM for embedding tables, (3) set up a loss function with tensor factors as loss function and perform SGD optimization. 1/3 to 1/2 page should be enough}
% \jiayi{didn't get the (3) set up a loss function with tensor factors as loss function and perform SGD optimization. }

Tensor decomposition has been widely adopted to \textcolor{blue}{train compressed neural networks from-scratch} \cite{yang2017tensor, ye2018learning,novikov2015tensorizing, hawkins2022towards,hawkins2021bayesian, ma2019tensorized}. Among the various low-rank tensor formats, TT and TTM are the most widely used due to their exceptional accuracy, orders of magnitude compression ratios, and straightforward data structures \cite{novikov2015tensorizing, wang2020compressing, wang2021nonlinear, yang2023quantization, yang2024comera}. In this work, we \textcolor{blue}{extend tensor-compressed training to transformer architectures}, primarily using TT to compress the weight matrices and TTM to compress the embedding table.

We consider the weight matirx $\mat{W}\in \mathbb{R}^{M\times N}$ in a linear layer. Assume $ {M=\prod_i^d {m_i}}$ and ${N=\prod_i^d {n_i}}$, and we reshape the matrix $\mat{W}$ into an order-$2d$ tensor $\ten{W}\in \mathbb{R}^{m_1 \times ... \times m_d\times n_1 \times ... \times n_d}$. With TT decomposition, we can represent $\ten{W}$ with TT cores $\{\ten{G}_k\}_{k=1}^{2d}$ as
\begin{align}
\ten{W} = \ten{G}_1\times_3^1...\ten{G}_k\times_3^1...\ten{G}_{d}\times_3^1...\ten{G}_{d+k}\times_3^1\ten{G}_{2d},
\label{eqa:TT-1}
\end{align}
Here $\ten{G}_k\in \mathbb{R}^{r_{k-1}\times m_{k} \times r_{k}}$ when $1\leq k\leq d$ and $\ten{G}_k\in \mathbb{R}^{r_{k-1}\times n_{k} \times r_{k}}$ when $d+1\leq k\leq 2d$.  
The total number of variables in $\ten{W}$ is reduced to $\sum_{k=1}^{d} (r_{k-1}m_kr_k + r_{k-1+d}n_kr_{k+d})$ from $\prod_{k=1}^d m_kn_k$, which approximately achieves $O(m^dn^d)\rightarrow O(dr^2(m+n))$ memory reduction.

For the embedding table, we follow \cite{yang2024comera,hrinchuk2020tensorized} and use TTM to compress the vocab dictionary. We first reshape the token embedding layer $\mat{E}_{\rm tok} \in \mathbb{R}^{M\times N}$ to $\ten{E}_{\rm tok}\in \mathbb{R}^{m_1 \times n_1... \times m_d\times n_d}$, where ${M=\prod_i^d {m_i}}$ and ${N=\prod_i^d {n_i}}$.
We define the TTM representation for the token embedding layer as 
\begin{align}
\ten{E}_{\rm tok} = \ten{F}_1\times_4^1...\ten{F}_k\times_4^1...\ten{F}_{d},
\label{eqa:TTM-1}
\end{align}
where $\ten{F}_k\in \mathbb{R}^{r_{k-1}\times m_{k} \times n_{k} \times r_{k}}$.
TTM reduces the number of parameters from $\prod_{k=1}^d m_kn_k$ to $\sum_{k=1}^{d} (r_{k-1}m_kn_kr_k)$, achieving $O(m^{d-1}n^{d-1}/dr^2)$ memory reduction.

\section{Tensorized Transformer Training}

\subsection{Overall Training Framework}
We consider the transformer in Fig. \ref{fig:Trans}, which has one embedding layer, $N$ encoder blocks and a task-specific classifier.
% We also elaborate the weight formats of each layer in both the uncompressed baseline and tensor-train Transformer. 
The three embedding tables for token, position, and segment are compressed to TTM as done in~\cite{yang2024comera}. All weight matrices (including those in the attention layers and feed-forward layers) of the encoder blocks are compressed to TT. The fully connected weight matrices in the classifier are also compressed to TT. The last task-specific linear layer for classification is kept uncompressed. In the training process, we use stochastic gradient descent (SGD) to solve the optimization problem
\begin{equation}
    \min \limits_{\boldsymbol{\theta}} {\cal L} (\boldsymbol{\theta}, {\cal D}),
\end{equation}
where ${\cal D}$ is the training dataset. Note that the training variables $\boldsymbol{\theta}$ include compression factors once a matrix is parameterized in a TT or TTM format. This differs from standard training where only large-scale 2-D matrices are trained. 

% \begin{figure}[htbp]
% \includegraphics[width=.5\textwidth]{figs/detail-ffn.png}
% \caption{Detail training phases of FP, BP, and WG of one feed-forward network (FFN) in encoder. The blue blocks denote the computing kernels for activation (Act) functions and different tensor contraction (TC) kernels are used for each training stage.The purple blocks are gradient for weights, while the white blocks are activations and gradient of activations.
% % \caption{Detail training phases of FP, BP, and WG of one feed-forward network (FFN) in encoder. The highlighted and gray blocks represent active and non-active data in the each phase. And the rounded blocks denote the computing kernels, including tensor contraction (TC) kernels and activation (Act) functions.
% }
% \label{fig:Detail}
% \end{figure}

Tensorized transformer training includes three stages: 
% \zz{provide high-level key idea here, and leave the details to other subsections.}
\begin{enumerate}
    \item {\bf Forward propagation (FP)}. 
    % \zz{Use a short paragraph and only one equation to describe what tensorized FWD do (i.e., given the tensor factors, perform layer-by-layer tensor contraction to do inference and get the loss. }
    In a forward pass, the input tokens are fed into the embedding layer to generate hidden features, then the $N$ encoders sequentially process the input features, and finally the classifier computes predictions and a loss function computes the loss using the predictions and labels. 
    The main computations, matrix-vector products, are replaced with low-rank tensor-network contractions, which leads to a significant reduction in memory and FLOPs.
    \item {\bf Backward propagation (BP)}.
    % \zz{Use a short paragraph and only one equation to describe what tensorized BWD do (i.e., given the tensor factors, perform layer-by-layer tensor contraction to get the gradient w.r.t. low-rank tensor factors.}
    The BP process first computes the gradients of the loss with respect to feature values according to a chain rule. After that, the gradient with respect to each TT factor is also computed via a tensor network contraction.
    \item {\bf Model parameter update (PU)}. The model parameters are updated as $\boldsymbol{\theta} \leftarrow \boldsymbol{\theta}  - \alpha \boldsymbol{\theta}'$, where $\alpha$ is a learning rate and $\boldsymbol{\theta}'=\frac{\partial {\cal L }}{\partial \boldsymbol{\theta}}$ is the gradient. Since we use tensor-compressed training, updates are also performed on TT or TTM factors for compressed layers. For example, a TT core $\ten{G}_k$ is updated as $\ten{G}_k \leftarrow \ten{G}_k - \alpha \ten{G'}_k$ in each iteration.
\end{enumerate}

\begin{figure*}[t]
\centering
\begin{align}
% \ten{G}'_{d+k}[i_{d+k}] = &
% \left( \sum_{j_t, t\neq k}
% \prod_{l=1,l\neq k}^d\ten{G}_{d+l}^{T}[j_t]x_{j_1,...j_d} \right)
% \left(\sum_{i_t}
% \prod_{l=1}^d\ten{G}_l^{T}[i_t]y'_{i_1,...,i_d}
% \right)
% \label{eqa:TTWG-1} \\
% \ten{G}'_k[i_k] = &
% \left( \sum_{j_t}
% \prod_{l=1}^d\ten{G}_{d+l}^{T}[j_{d+t}]x_{j_1,...j_d} \right)
% \left( \sum_{i_t, t\neq k}
% \prod_{l=1, l\neq k}^d\ten{G}_{l}^{T}[i_t]y'_{i_1,...,i_d} \right)
\ten{G}'_{d+k}[i_{d+k}] = &
\sum_{i_1,...,i_d, j_1, ...,j_{k-1},j_{k+1},...j_{d}}
\left( \ten{G}_{d+k-1}^{T}[j_{k-1}]...\ten{G}_1^{T}[i_1]y'_{i_1,...,i_d} \right)
\left( x_{j_1,...j_d} \ten{G}_{2d}^{T}[j_{d}]...\ten{G}_{d+k+1}^{T}[j_{k+1}]\right), 
\label{eqa:TTWG-1} \\
\ten{G}'_k[i_k] = &
\sum_{i_1,...,i_{k-1},i_{k+1},...,i_{d}, j_1,...,j_d}
\left( \ten{G}_{k-1}^{T}[i_{k-1}]...\ten{G}_{1}^{T}[i_1]y'_{i_1,...,i_d} \right)
\left( 
x_{j_1,...j_d}\ten{G}_{2d}^{T}[j_{d}]...\ten{G}_{k+1}^{T}[i_{k+1}] \right), \quad {\rm for}\; k\in [1,d] 
\label{eqa:TTWG-2} \\
\ten{F}'_k[i_k,j_k] = &
\sum_{i_1,...,i_{k-1},i_{k+1},...,i_{d}}
\left( \ten{F}_{k-1}^{T}[i_{k-1},j_{k-1}] ... \ten{F}_{1}^{T}[i_{1},j_{1}] y'_{i_1,...,i_d} \right)
\left(\ten{F}_{d}^{T}[i_{d},j_{d}]...\ten{F}_{k+1}^{T}[i_{k+1},j_{k+1}] \right)
\label{eqa:TTM-WG}
\end{align}
\hrulefill
\end{figure*}

% \begin{figure*}[t]
%     \centering
%     \subfloat[TT-FP (d=2)]{%
%         \includegraphics[width=0.3\textwidth]{fig_up/TT-FP.png} % Replace with your image
%         \label{fig:TT-FP}
%     }
%     % \hfill
%     \subfloat[TT-BP (activation gradient)]{%
%         \includegraphics[width=0.3\textwidth]{fig_up/TT-BP.png} % Replace with your image
%         \label{fig:TT-BP}
%     }
%     % \hfill
%     \subfloat[TT-BP (TT gradients)]{%
%         \includegraphics[width=0.3\textwidth]{fig_up/TT-WG.png} % Replace with your image
%         \label{fig:TT-WG}
%     }
%     \caption{Tensor graph representations for TT- FP and BP of one linear layer. Here Node $0$ represent input tensor $\ten{X}$ of a linear layer. The purple nodes $1\sim 4$ denote TT cores $\ten{G}_1 \sim \ten{G}_4$. Node $5'$ denotes the gradient tensor $\ten{Y}'$. In (c), we only show the tensor network for computing the gradient with respect to $\ten{G}_4$.}
%     \label{fig:TG}
% \end{figure*}

\begin{figure*}[t]
    \centering
        \includegraphics[width=0.8\textwidth]{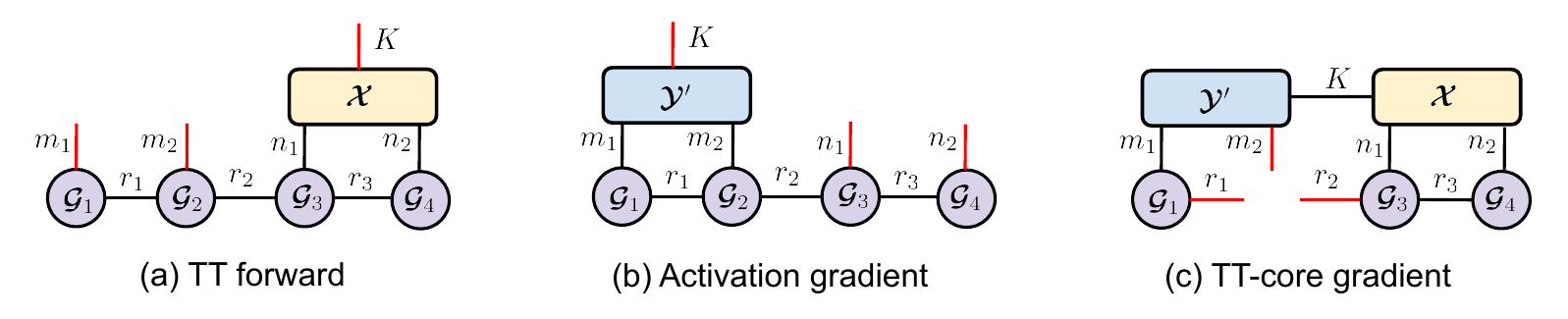} 
    \caption{Tensor graph representations for the TT-format (a) forward propagation, (b) gradient computation w.r.t. activations and (c) gradient computation w.r.t. a TT core (with $\ten{G}_2$ as an example) of a linear layer. For simplicity, here we assume $d=2$ and that the matrix is fold to a tensor of order $2d=4$. The red lines denote free edges of a tensor network, which indicate the dimension of the result tensor.}
    \label{fig:TG}
\end{figure*}

\subsection{Tensorized Linear Layer}

We use tensorized linear layers in most computing blocks of a transformer, including the feed-forward networks and the linear transforms in the attention layers.

{\bf Forward propagation.}
\label{subsec:fwd}
% \zz{Shorten this subsection, and use a tensor network to describe the FWD process of each layer. }
Most computations in the forward propagation of a transformer involve the matrix-vector product $\mat{y}=\mat{W} \mat{x}$, with $\mat{W}\in \mathbb{R}^{M\times N}$ and $\mat{x} \in \mathbb{R}^{N}$. Assume that $M=\prod\limits_{i=1}^d m_i$ and $N=\prod \limits_{i=1}^d n_i$, and that $\mat{W}$ is reshaped to a tensor $\ten{W}$ of order $2d$. With the TT representation in \eqref{eqa:TT-1}, the matrix-vector product is efficiently computed by low-rank tensor network contraction in our framework.   
Given the $2d$ TT cores $\{ \ten{G}_k\}_{k=1}^{2d} $ that represent $\mat{W}$, we compute $\mat{y}=\mat{W}\mat{x}$ in a tensor-compressed form as
\begin{equation}
\begin{aligned}
&y_{i_1,...i_d} = \ten{G}_{1}[i_1]...\ten{G}_{d}[i_d]\sum_{j_1,...,j_d}\ten{G}_{d+1}[j_1]...\ten{G}_{2d}[j_d]x_{j_1,...,j_d}
\label{eqa:TT-FP}
\end{aligned}
\end{equation}
where $x_{j_1,...,j_d}$ and $y_{i_1,...i_d}$ are elements in the tensor-format input $\ten{X} \in \mathbb{R}^{n_1\times n_2\times \cdots n_d}$ and tensor-format output $\ten{Y} \in \mathbb{R}^{m_1\times m_2\times \cdots m_d}$, separately. Here $\ten{G}_k [i_k]$ denotes one slice of the order-$3$ TT core $\ten{G}_k$ by fixing its $2$nd index as $i_k$. The above equation shows that contraction happens among $\ten{X}$ and the last $d$ TT cores. 
Fig.~\ref{fig:TG}(a) shows the graph representation of the above equation when $d$ is $3$. Here, we have an additional free edge of size $K$, because in practice we often perform training with an input dimension $K$ which is the multiplication of a batch size multiplied with the sequence length. Therefore, $\mat{x}$ should be replaced with a matrix $\mat{X} \in \mathbb{R}^{N\times K}$. 

% % According to the above equations for TTM- and TT- FP, to compute each element y, we need to accumulate the results of d and 2d matrix-multiplications between a r-by-r slice of a tensor core and a r-by-1 intermediate result, which is $O(dr^2)$ and $O(2dr^2)$ complexity. Therefore, for the whole output vector, $O(m^{d}n^{d}dr^2)$ and $O((m^{d}+n^{d})dr^2)$ complexity are required, respectively.
% % Furthermore, in the TNNs, inputs of the linear layers are always B-by-M matrices, where B represents the batch-size (or sequence length in NLP tasks), let $m1 = ... = m_d = n_1 = ... = n_d, r_1 = ... = r_{d}$, the time complexity could be extended to $O(Bn^{2d}dr^2)$ and $O(2Bn^{d}dr^2)$.
% % However, the element-by-element computational process of tensorized neural networks is extremely inefficient and might even have greater time complexity of uncompressed networks. Therefore, some previous works \cite{novikov2015tensorizing, deng2019tie} proposed to transfer the computing flow to matrix-matrix multiplications to reduce the time complexity for TTM-format inference, and we extend the computing flow for TTM- and TT- format training.
% \zz{Do NOT talk about the contractions sequence here. Talk about the contraction sequence in Section IV. }

% % as shown in Fig. \ref{fig:TTM}, 
% % in the k-th step, 
% $B\prod_{i=1}^{d-k+1}n_i \prod_{i=d-k+1}^{d} m_{i} r_{d-k}r_{d-k+1}$ multiplications are required, and thus the total computational cost of d-step TTM-FP as: 

% % $O(Bn^{d+1}dr^2))$.

{\bf Backward propagation.}
\label{subsec: BP}
% \zz{Also discribe the computing task via a tensor network, and ignore the contraction sequence, which could be deferred to Section IV. }
% In MM-format BP, the 2-d weight matrix $\mat{W}\in \mathbb{R}^{M\times N}$ is multiplied by the output gradient $\nabla_L\mat{Y}\in \mathbb{R}^{B\times M}$ and results in $\nabla_L\mat{X}\in \mathbb{R}^{B\times N}$. 
With low-rank TT parameterization, the BP process for gradient computation can also be done via tensor network contractions. Let us first define the following two tensors: 
\begin{equation}
    \ten{X}'=\frac{\partial {\cal L}}{\partial \ten{X}} , \quad \ten{Y}'=\frac{\partial {\cal L}}{\partial \ten{Y}}, 
\end{equation}
which are the order-$d$ tensor representations for $\mat{x}'=\frac{\partial {\cal L}}{\partial \mat{x}}$ and $\mat{y}'=\frac{\partial {\cal L}}{\partial \mat{y}}$, respectively. In standard uncompressed training with $\mat{y}=\mat{W}\mat{x}$, it is known that
\begin{equation}
    \mat{W}'=\frac{\partial {\cal L}}{\partial \mat{W}}=\mat{y}' \mat{x}^T. 
\end{equation}

With $\ten{Y}'$ being the high-order tensor representation for $\mat{Y}'$, the gradient with respect to the TT factors can be computed in two steps:
\begin{itemize}[leftmargin=*]
    \item We first compute $\ten{X}'$ via a tensor-network contraction
\begin{equation}
x'_{j_1,...j_d} = 
\ten{G}_{2d}^{T}[j_d]...\ten{G}_{d+1}^{T}[j_1]\sum_{i_1,...,i_d}\ten{G}_{d}^{T}[i_d]...\ten{G}_{1}^{T}[i_1]y'_{i_1,...,i_d}. 
\label{eqa:TT-BP}
\end{equation} \normalsize
\item Then the gradients with respect to the $2d$ TT cores can be computed according to Eq. \eqref{eqa:TTWG-1} and Eq. \eqref{eqa:TTWG-2} for $k \in [1,d]$. Note that $\ten{G}'_k[i_k]$ denotes the $i_k$-th slice of the derivative tensor $\ten{G}_k'=\frac{\partial {\cal L}}{\partial \ten{G}_k}$ by fixing its 2nd index as $i_k$. 
\end{itemize}
Fig.~\ref{fig:TG} (b) and Fig.~\ref{fig:TG} (c)  show the tensor networks of activation gradient and TT core
gradients in one linear layer. Note that Fig.~\ref{fig:TG} (c) illustrates how to compute $\ten{G}'_4$ by eliminating $\ten{G}_4$ from the computing graph. In general, we can remove $\ten{G}_k$ from the tensor network while keeping all other nodes if we want to compute the gradient $\frac{\partial {\cal L}}{\partial \ten{G}_k}$.

\subsection{Tensorized Embedding Table}
The embedding layer in a transformer has huge look-up tables, which consumes a large amount of memory. The TTM representation in \eqref{eq: ttm} can greatly reduce the memory cost of embedding layers, but its forward and backward propagation uses different tensor-network contractions. 

{\bf Forward propagation.} Given the $d$ TTM factors $\{ \ten{F}_k\}_{k=1}^{d} $ that represent $\rm{E}_{tok} \in \mathbb{R}^{M\times N}$, where $M=\prod\limits_{i=1}^d m_i$ and $N=\prod \limits_{i=1}^d n_i$, and thus $\rm{E}_{tok}$ is reshaped to a tensor $\ten{E}$ of order $2d$. In the look-up process, a column of $\rm{E}_{tok}$ is selected for each input token. We use an index vector $\mat{j} =[j_1, j_2, \cdots j_d]$ to select the $j_k$ -th subtensor of each TTM core $\ten{F}_k$.
The output feature $\mat{y}$ could be computed as
% \begin{align}
% &y_{i_1,...i_d} = \sum_{j_1,...,j_d}\ten{F}_1[i_1,j_1]...\ten{F}_d[i_d,j_d]x_{j_1,...,j_d},
% \label{eqa:TTM-FP}
% \end{align}
\begin{align}
&y_{i_1,...i_d} = \ten{F}_1[i_1,j_1] \ten{F}_2[i_2,j_2]...\ten{F}_d[i_d,j_d],
\label{eqa:TTM-FP}
\end{align}
where $\ten{F}_k [i_k,j_k]$ is a $2$-D slice of $\ten{F}$ by fixing its $2$nd and $3$rd indices as $i_k$ and $j_k$, respectively. 
According to the above equation, using the TTM embedding table, we first look up each TTM core using the $d$-element index $\mat{j}$, and use the selected slices to form the output feature.
% In TTM-format, $\ten{X}$ needs to contract with all TTM cores, and the graph representation is shown in Fig.~\ref{fig:TG} \zz{to be updated} for batch size $K$.  

{\bf Backward propagation.}
% We first compute $\ten{X}'$ via a tensor-network contraction
% \begin{equation}
% \begin{aligned}
% x'_{j_1,...j_d} = \sum_{i_1,...,i_d}\ten{F}_d^{T}[i_d,j_d]...\ten{F}_1^{T}[i_1,j_1]y'_{i_1,...,i_d}.
% \label{eqa:TTM-BP}
% \end{aligned}
% \end{equation}
Since the input index does not require gradients, we only compute the gradients for the tensor cores in the TTM embedding table according to Eq.~\eqref{eqa:TTM-WG}. Here, $\ten{F}'_k[i_k,j_k]$ denotes a slice of the derivative tensor $\ten{F}_k'=\frac{\partial {\cal L}}{\partial \ten{F}_k}$ by fixing its $2$nd and $3$rd indices as $i_k$ and $j_k$, respectively. 

\section{Algorithm Implementation}
\label{AA}

As discussed earlier, training of tensor-compressed models is based on tensor contractions. 
\textcolor{blue}{While the contraction order does not affect the training curve, the associated computational and memory complexities are highly sensitive to it.} 
Existing tensor-compressed inference accelerators typically adopt a right-to-left contraction sequence for both tensor-train matrix (TTM) \cite{novikov2015tensorizing, deng2019tie} and TT formats \cite{gong2023ette}.
In this section, we propose an optimized contraction order for the tensor-train (TT)-format linear layer to \textcolor{blue}{enhance the computational and memory efficiency without changing the model performance.}

\begin{figure*}[t]
\centering
% \begin{equation}
% \begin{aligned}
% \ten{F}'_k[i_k,j_k] = \sum_{i_1,...,i_{k-1},i_{k+1},...,i_{d}}
% &\left( \ten{F}_{k-1}^{T}[i_{k-1},j_{k-1}] ... \ten{F}_{1}^{T}[i_{1},j_{1}] y'_{i_1,...,i_d} \right)
% \left(\ten{F}_{d}^{T}[i_{d},j_{d}]...\ten{F}_{k+1}^{T}[i_{k+1},j_{k+1}] \right)
% % \ten{F}'_k[i_k,j_k] = \sum_{i_1,...,i_{k-1},i_{k+1},...,i_{d}, j_1, ...,j_{k-1},j_{k+1},...j_{d}}
% % &\left( \ten{F}_{k-1}^{T}[i_{k-1},j_{k-1}] ... \ten{F}_{1}^{T}[i_{1},j_{1}] y'_{i_1,...,i_d} \right)
% % \left(x_{j_1,...j_d}\ten{F}_{d}^{T}[i_{d},j_{d}]...\ten{F}_{k+1}^{T}[i_{k+1},j_{k+1}] \right)
% \label{eqa:TTM-WG}
% \end{aligned}
% \end{equation}
% \hrulefill

\begin{align}
{\rm MUL}_{\rm TT} &= K\sum_{k=0}^{d-1}\left(
r_{2d-k-1}r_{2d-k}\prod_{i=1}^{d-k}n_i
 + 
r_{d-k-1}r_{d-k}\prod_{i=d-k}^{d}m_{i}
\right) \label{eqa:comp-TT} \\
{\rm Memory}_{\rm TT} &= Kr_d+
K\sum_{k=0}^{d-2}\left(
 r_{2d-k-1}\prod_{i=1}^{d-k-1}n_i +
 r_{d-k-1}\prod_{i=d-k}^{d}m_{i}\right) \label{eqa:mem-TT}
\end{align}\\
\hrulefill
\end{figure*}
\begin{figure*}[t]
\centering
\begin{align}
\begin{split}\label{eqa:comp-BTT}
{\rm MUL}_{\rm BTT} &=  \sum_{k=0}^{d-2}
\left(
r_{2d-k-1}r_{2d-k-2}\left(\prod_{i=d-k-1}^{d} n_{i}\right)  +
r_{k+1}r_{k+2}\left(\prod_{i=1}^{k+2} m_i\right) 
\right)
+ Kr_d\left(\prod_{i=1}^{d}m_i + \prod_{i=1}^{d}n_i\right)
\end{split}\\
\hrulefill
{\rm Memory}_{\rm BTT} &=  
\sum_{k=0}^{d-2}\left(r_{2d-k-2}\left(\prod_{i=d-k-1}^{d} n_{i}\right) + r_{k+1}\left(\prod_{i=1}^{k+2} m_i\right)\right)  + Kr_d.
\label{eqa:mem-BTT} 
\end{align}
\hrulefill
\end{figure*}

\subsection{Complexity of Right-to-Left Contractions}
%Given $\mat{W}\in \mathbb{R}^{M\times N}$ and $\mat{X}\in \mathbb{R}^{N\times K}$, the matrix-matrix product $\mat{W} \mat{X}$ has a computing complexity of $O(KMN)$ in standard neural network training.

We first evaluate the computational and memory complexity of the right-to-left TT contraction in a linear layer. We only analyze the forward propagation, since computing the gradients with respect to activations and TT cores uses a similar contraction process.
% since the dimensions of computing dimensions in tensor contraction (or matrix multiplication) involved in different training stages are the same with only the positions of dimensions exchanged. 
% Specifically, Fig. \ref{fig:TCseq} shows the shapes of activation and weights involved in each computation, 
The total computational cost in training is roughly $3\times$ of the inference cost.

% We extend the computing flow for training in Fig. \ref{fig:TCseq} and analyze the computational complexity below.

{\bf Computational Cost.} The forward of a TT-format linear layer has $2d$ contraction steps. \textcolor{blue}{The total number of multiplications in a forward pass is estimated in Eq. \eqref{eqa:comp-TT}, where the first and second terms accounts for the number of multiplications in the initial and final $d$ contraction steps, separately.}
% Generally, the $k$-th contraction step (with $0 \leq k\leq d-1)$ involves $Kr_{2d-k-1}r_{2d-k}\prod_{i=1}^{d-k}n_i $ multiplications; the $(d+k)$-th step involves $Kr_{d-k-1}r_{d-k} \prod_{i=d-k}^{d}m_{i} $ multiplications. As a result, the total number of multiplications in a forward pass is estimated in Eq. \eqref{eqa:comp-TT}. 

{\bf Memory Cost.} Training with the TT-format incurs additional intra-layer memory overhead due to the multiple contraction steps involved, which is estimated in Eq. \eqref{eqa:mem-TT}. 
\textcolor{blue}{Specifically, there are $2d-1$ intermediate results generated during $2d$ contraction steps, , excluding the final step which produces the output and is therefore not counted. 
The first term in the equation captures the memory cost at the 
$d$-th contraction step, while the second and third terms correspond to the costs incurred during the initial and final $d-1$ steps, respectively.}
% Generally, the $k$-th contraction step (with $0 \leq k< d-1)$ requires $Kr_{2d-k-1}\prod_{i=1}^{d-k-1}n_i $ memory cost; the $(d+k)$-th step requires $Kr_{d-k-1} \prod_{i=d-k}^{d}m_{i} $ memory. 
All of these intermediate results need to be stored for reuse in back propagation. 
% As a result, the total number of multiplications in a forward pass is estimated in Eq. \eqref{eqa:comp-TT}. 
% with the computational complexity as $O(Bn^{d+1}((d-2)r^2 + 2r))$, considering $r_0 = r_d = 1$.
% \begin{figure*}[t]
% \centering
% \begin{align}
% \begin{split}\label{eqa:mem-TT}
% &MEM_{total} (TT) =
% B\sum_{k=0}^{d-2}\left(
%  r_{2d-k-1}\left(\prod_{i=1}^{d-k-1}n_i\right) +
%  r_{d-k-1}\left(\prod_{i=d-k}^{d}m_{i}\right)\right) + Br_d.
% \end{split}\\
% \end{align}
% \hrulefill
% \end{figure*}
% The intra-layer memory consumption is also a critical property for hardware efficiency, which corresponds to the intra-layer activation generated from FP and BP and resued in WG, and could be formulated as 

\begin{figure}[t]
     \centering
     \includegraphics[width=.5\textwidth]{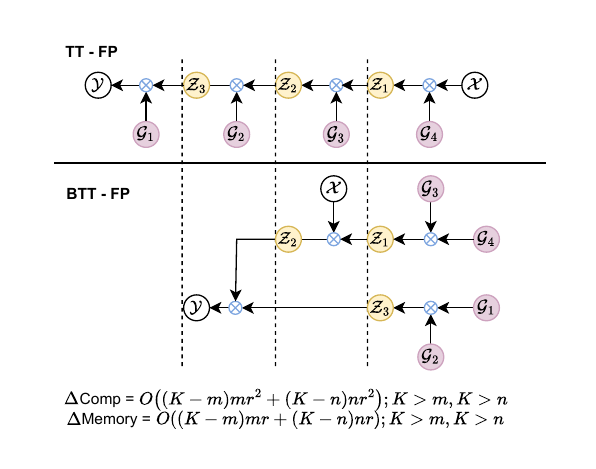}
    \caption{Comparison of the computing flow of the TT-format and our modified BTT forward propagation when $d=2$. Contraction operations are represented in blue multipliers. Here white nodes represents input tensor $\ten{X}$ and output tensor $\ten{Y}$; purple nodes are TT cores $\ten{G}_1$ to $\ten{G}_4$; nodes marked in yellow denotes the intermediate contraction results of the involved tensors.}
    % \caption{Comparison of the computing flow of MM-, TTM-, TT and our modified BTT- FP, BP, and WG. Shapes of activation and weights are annotated inside the blocks. Contraction operations in FP, BP and WG are represented in black, red, and blue multipliers. And arrows with different colors connected with activation blocks notes different values in each stage.}
     \label{fig:TCseq}
\end{figure}

Fig.~\ref{fig:TCseq} (top) shows the contraction sequence for a linear layer in TT format when $d=2$. In the right-to-left contraction flow, the computation and memory complexity of every step depend on the input dimension $K$ which is the batch size multiplied by the sequence length. In NLP and computer vision tasks, $K$ can be large, leading to low hardware efficiency.
To address this limitation, we propose an efficient bidirectional contraction scheme that reduces computational and memory costs by eliminating the dependence on $K$ in most contraction steps.

\subsection{Bi-directional Tensor-Train Contraction}
\label{sec:BTT}
Our bidirectional tensor-train (BTT) contraction is shown in Fig.~\ref{fig:TCseq} (bottom). 
Since the left and right $d$ TT cores have no data dependency, we can perform contraction from the left and right towards the middle in parallel. 
This can reduce the total number of computation stages from $2d$ to $d+1$. 
\textcolor{blue}{To clearly demonstrate the efficiency brought by our BTT contraction, we formulate the computing and memory costs and conduct a comprehensive comparison with the right-to-left TT contraction.}
% To better explain the efficiency of the training, we will calculate the exact computation and memory costs for each training stage and compare it with the original right-to-left TT contraction.

\begin{figure}[t]
     \centering
     \includegraphics[width=.48\textwidth]{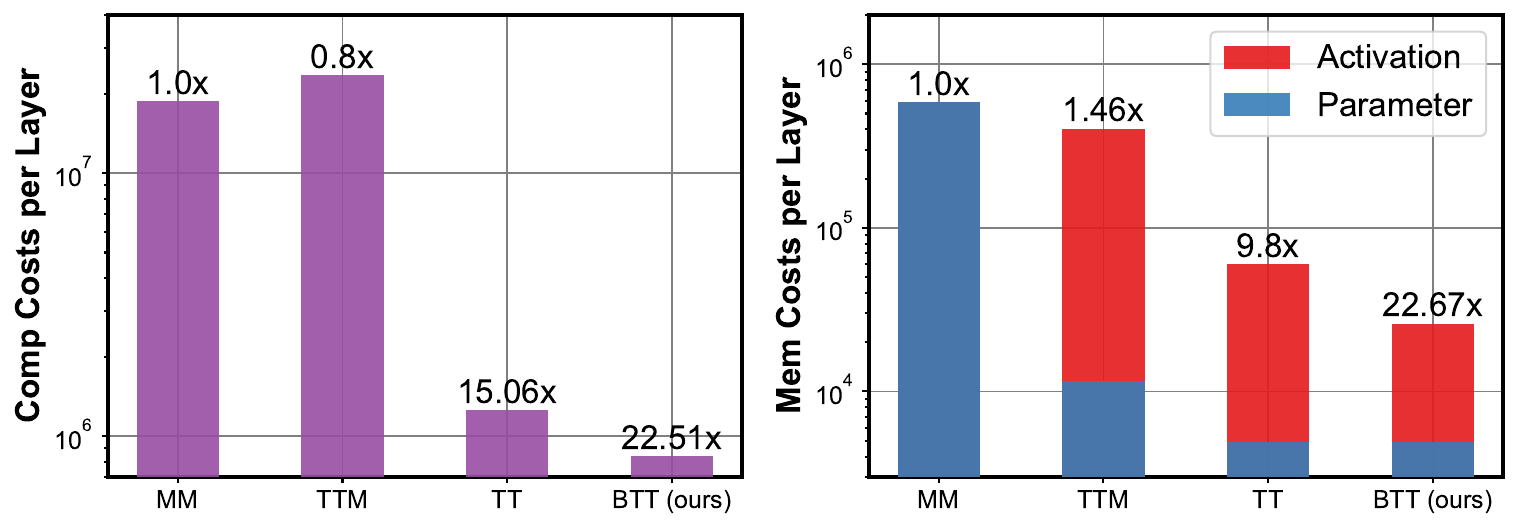}
    \caption{Comparison of the computation and memory costs of MM, TTM, TT and our BTT.}
    \label{fig:speedup}
\end{figure}

% Although the total number of contractions (denoted as blue multipliers in Fig.~\ref{fig:TCseq}) remains unchanged, the computational cost can be significantly reduced by eliminating the dependence on $K$ in the first $d-1$ contraction steps.
\textcolor{blue}{The core idea of our BTT is to modify the computing path of the tensor-compressed linear layer to explore the parallelism and reduce involving large $K$ in most contraction steps to reduce the computing and memory efficiency. 
}
% In the $k$-th step, we perform a contraction between a matrix of size $\left(\prod_{i=1}^{k}m_i \right) \times r_k$ and the TT core $\ten{G}_k \in \mathbb{R}^{r_k\times m_{k+1}\times r_{k+1}}$ on the left, as well as the contraction between the TT core $\ten{G}_{2d-k} \in \mathbb{R}^{r_{2d-k-1}\times n_{d-k}\times r_{2d-k}}$ and a matrix of size $r_{2d-k} \times \prod_{i=1}^{k}n_{d-i+1}$ on the right.
The resulting total computational cost and memory consumption are estimated as Eq. \eqref{eqa:comp-BTT} and Eq. \eqref{eqa:mem-BTT}. \textcolor{blue}{In both equations, the first term—independent of the factor $K$—captures the costs associated with contracting the left and right TT cores during the initial $d$ steps. In contrast, the final term, which is scaled by $K$, reflects the costs incurred in the final two contraction steps.}
% Note that the first terms in both equations are independent of $K$. 
Based on the above equations, BTT always have lower computational complexity when $m_i$ and $n_i$ are smaller than $K$, which is common in modern NLP and CV tasks.
\begin{table*}[t]
\centering
\caption{Computational and memory training complexities of each linear layer for TTM, right-to-left TT contraction, and the proposed BTT contraction method.}
\begin{tabular}{|c|c|cc|}
\hline
\multirow{2}{*}{\textbf{Method} }    & \multirow{2}{*}{\textbf{FLOPs}} & \multicolumn{2}{c|}{\textbf{Memory}}              \\ \cline{3-4} 
                          &                              & \multicolumn{1}{c|}{\textbf{Weight}} & \textbf{Activation} \\ \hline
Matrix-Matrix multiplication (MM)                      &  $O(3Kn^{2d})$               & \multicolumn{1}{c|}{$O(n^{2d})$}       & 0           \\ \hline
Tensor-Train-Matrix contraction (TTM)                      &  $O(3Kn^{d+1}((d-2)r^2 + 2r))$                           & \multicolumn{1}{c|}{$O(n^2((d-2)r^2+2r))$}       & $O(Kn^d(d-1)r)$           \\ \hline
Right-to-left Tensor-Train contraction (TT)                        &  $O(6K(\sum \limits_{k=1}^{d-1} n^k r^2 + n^dr))$             & \multicolumn{1}{c|}{\multirow{2}{*}{$O(2n((d-2)r^2+2r))$}}    &  $O(2K\sum \limits_{k=1}^{d-1}n^{k}r + Kr)$            \\ \cline{1-2} \cline{4-4} 
\textbf{Bidirectional Tensor-Train contraction (BTT, ours)}       &  $O(6\sum \limits_{k=2}^{d}n^{k}r^2+6Kn^dr)$                            & \multicolumn{1}{c|}{}      & $O(2\sum \limits _{k=2}^{d}n^{k}r + Kr)$         \\ \hline %\cline{1-2}
% \textbf{PBTT (ours)}      &  $O(3\sum \limits_{k=2}^{d}n^{k}r^2+6Kn^dr)$                            &    \multicolumn{1}{c|}{}       &            \\ \hline
% \textbf{PBTT - TT (Avg.)} &   $O(3(n-2K)\sum \limits_{k=1}^{d-1}n^{k}r^2)$                           & \multicolumn{1}{c|}{0}       &  $O(2(n-K)\sum \limits_{k=1}^{d-1}n^{k}r)$             \\ \hline
\end{tabular}
\label{table:complex}
\end{table*}

% The total computational costs for tensor linear layer training is three times of the tensor inference cost. 
To provide a comprehensive comparison between our BTT method and existing contraction schemes, we summarize the theoretical computing and memory costs of various methods in Table \ref{table:complex}. 
% The details complexity analysis for TTM is presented in Appendix \ref{sec:appendix}. 
For simplicity, we assume $m = n$ in all equations. 

% For the parallel BTT, the constraint is even relaxed to to $n<2K$.

{\bf Example.} We consider a hidden dimension of $768$ for the query matrix, and set $d=3$ with $\{n_1,n_2,n_3\}$ and $\{m_1,m_2,m_3\}$ being $\{12,8,8\}$ and $\{8,8,12\}$, respectively. We also set the TT rank to $12$ and the sequence length to $32$. 
Fig. \ref{fig:speedup}  compares the computation and memory costs of a linear layer implemented with different contraction methods. Clearly, our BTT contraction method consumes the least computation and memory cost. Compared with standard matrix-matrix multiplication, our BTT method is $22.51\times$ more computing efficient and $22.67\times$ more memory efficient. Compared with the right-to-left TT contraction, our BTT implementation further reduces the computing cost and the memory cost by $1.49\times$ and $2.31\times$, respectively.

\begin{figure}[t]
     \centering
     \includegraphics[width=.48\textwidth]{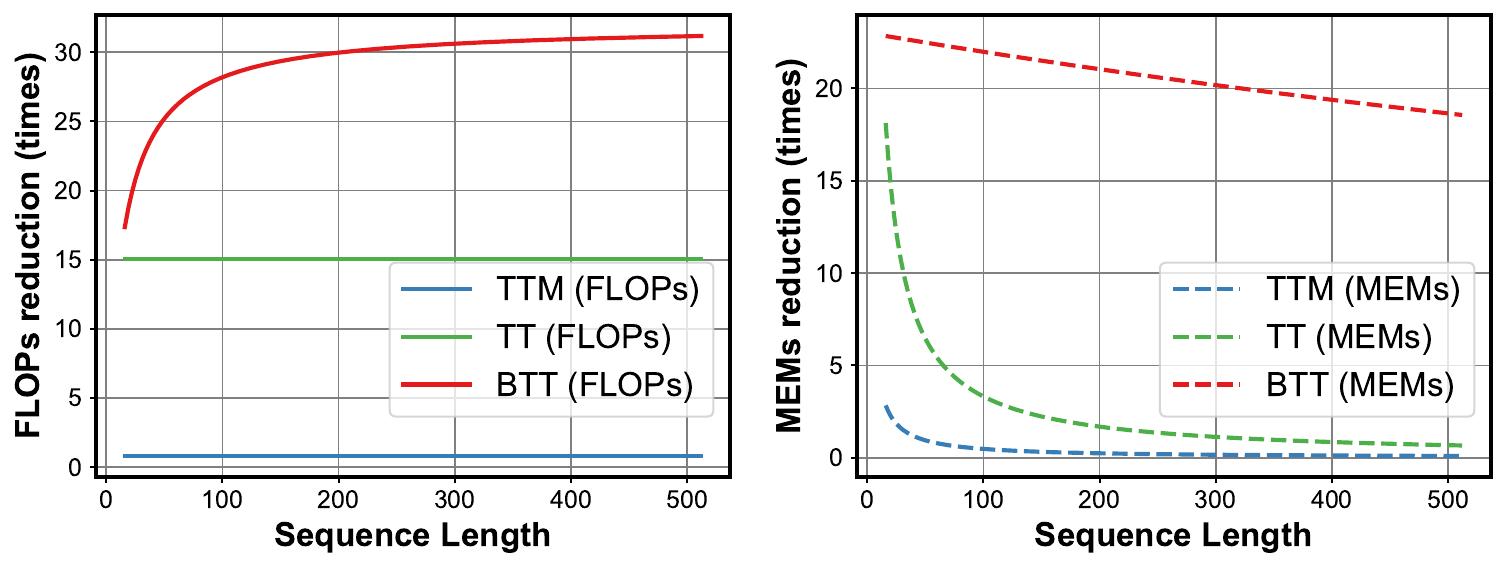}
     \includegraphics[width=.48\textwidth]{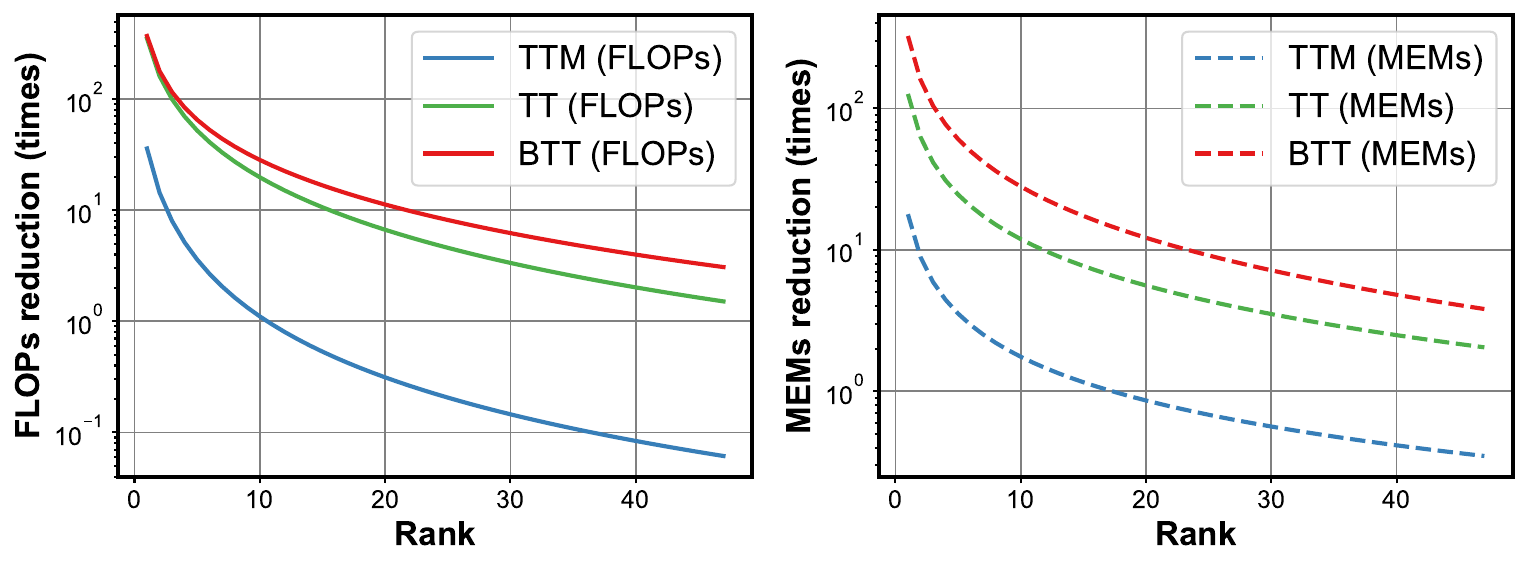}
    \caption{Computational and memory costs of TTM-based contraction, TT-based contraction and our BTT-based contraction corresponding to different sequence length (top) and rank (bottom). The reduction ratio is calculated based on matrix-matrix multiplication.}
    \label{fig:scalespeedup}
\end{figure}

% \subsubsection{Dependence on TT Ranks and Sequence Length}
\textcolor{blue}{To evaluate the influences of the hyper-parameters on our BTT contraction scheme, we show the computing and memory costs associated with different sequence length and TT ranks in Fig.~\ref{fig:scalespeedup}. The standard matrix-matrix multiplication is used a baseline to estimate the reduction ratios regarding FLOPs and memory cost.
In the upper figure, we fix the TT rank as $12$ and increase the sequence length from $8$ to $512$. In the bottom figure, we fix the sequence length as $32$ and increase the TT rank from $1$ to $48$.
As the sequence length increases, our BTT contraction shows greater advantages over the right-to-left contraction of TT and TTM methods in terms of both FLOPS and memory cost.
As the TT rank increases, the computing and memory advantage of all tensor-compressed methods degrades, but our BTT scheme always has the highest reduction ratios in both FLOPs and memory costs. }%Additionally, since larger rank and sequence length are required for a stronger language model to solve more complex tasks, our bi-directional TT contraction method could be more efficient for complex tasks on resource-constrained platforms and makes large language model training on edge possible.

% \begin{figure}[htbp]
% \includegraphics[width=.5\textwidth]{figs/workflow.png}
% \caption{Workflow during each training iteration. Two PE arrays are used for tensor contraction in each training stage for intra-stage parallelism. Purple blocks represent weights and gradients of weights stored in BRAM; Yellow blocks represent activation and gradient of activation stored in BRAM; Gray block represents off-chip memory.}
% \label{fig:workflow}
% \end{figure}

\section{FPGA Accelerator Design}
Based on the new proposed computing scheme, we develop an FPGA accelerator for tensor-compressed training.
\textcolor{blue}{While the BTT contraction strategy effectively mitigates the sequential bottleneck inherent in conventional tensor contraction and significantly reduces both FLOPs and memory usage, it introduces several trade-offs—such as increased resource consumption due to duplicated compute and memory buffers, and elevated control complexity within the FPGA implementation.
To address these challenges, we propose two key hardware-aware design strategies. First, we apply task rescheduling and tensor fusion to improve compute and memory efficiency during tensor-compressed transformer training. Second, we propose an on-chip memory management scheme based on tensor grouping, which enables efficient storage and access of tensor-compressed weights.}
% We perform hardware-level optimization to further reduce latency and memory usage. We also employ intra-stage and inter-stage dataflow optimization to reduce runtime and memory cost.
Our training accelerator is implemented in C++ and transferred to RTL via high-level synthesis (HLS). 

% CU design overview for each training stage
\subsection{Overall Architecture}
\label{sec:workflow}

\begin{figure*}[t]
    \centering
    \subfloat[]{%
        \includegraphics[width=0.5\textwidth]{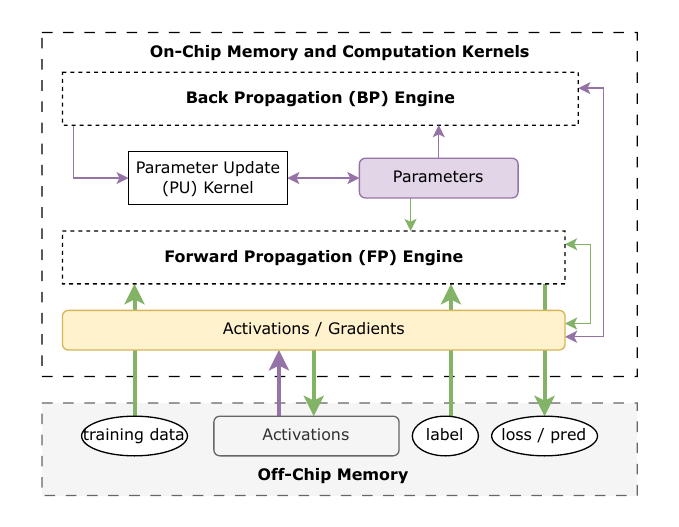} % Replace with your image
        \label{fig:arch-over}
    }
    \hfill
    \begin{minipage}[b]{0.44\textwidth}
    \subfloat[]{%
        \includegraphics[width=\textwidth]{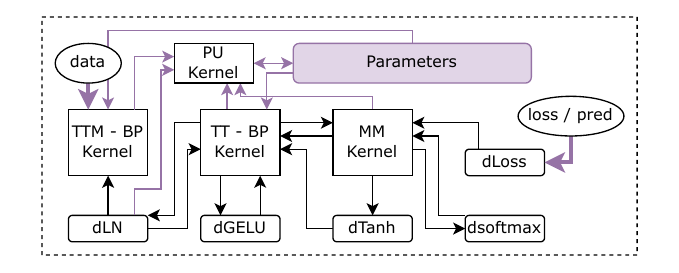} % Replace with your image
        \label{fig:arch-bp}
    } \\
    % \hfill
    \subfloat[]{%
        \includegraphics[width=\textwidth]{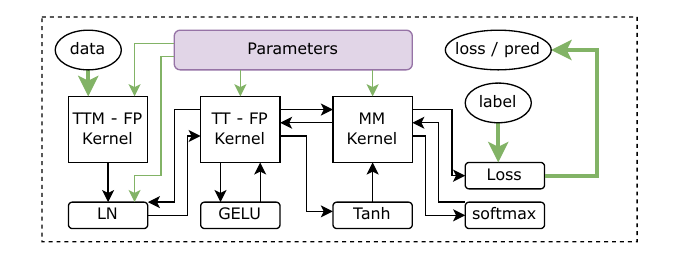} % Replace with your image
        \label{fig:arch-fp}
    }
    \end{minipage}
    \caption{(a): Overall architecture of the tensorized transformer training accelerator. (b) and (c): The detailed interactions between bottom-level computing kernels during backward and forward propagation. Note that in (b) and (c), the on-chip memory storing activation / gradient interacts with all computing kernels and thus is not shown in the figures.}
    \label{fig:arch}
\end{figure*}

Fig.~\ref{fig:arch} (a) shows the overall architecture of our tensor-compressed transformer training accelerator, including tensor-compressed forward propagation (FP) and backward propagation (BP) engines, on-chip and off-chip storage, and data transfer between difference training stages.

Due to the small number of model parameters caused by tensor compression, all training parameters and most activation values are stored in on-chip memory. This reduces the latency and energy overhead caused by off-chip communications. 
Since the volume of model parameters and activation values increase linearly with the number of layers, the total memory cost still exceeds the on-chip memory budget, thus off-chip memory access is inevitable.
We observe that the activation values consume much more memory than the tensor-compressed model parameters. Some interlayer activations produced in FP are only needed in BP. To support more complex tasks with more layers, we store these activations in off-chip memory and fetch them to on-chip memory when computing the gradients with respect to tensor factors.
% Since the read and write operations of activations can be done simultaneously with other computing kernels without data dependency, which eliminates the latency overhead. 

% Referring to Fig. \ref{fig:TCseq}, there exists intra-stage parallelism in FP and WG in BTT which could lead to latency reduction, therefore, two computing units (CU) are used for each training stage to achieve the intra- and inter- stage parallelism, each consists a 1-d processing element (PE) array.
The detailed computing kernels and dataflows inside the tensor-compressed BP and FP engines are shown in Figs.~\ref{fig:arch} (b) and (c), respectively.
To support the whole training progress, multiple computing kernels are developed for various computing schemes, including tensor contraction in the TTM-format embedding table and TT-format linear layers, matrix-multiplication (MM) in the attention parts and classifier, and various non-linear functions (softmax, GELU, Tanh, LayerNorm) used in transformer models. 

\subsection{Efficient Parallelism for Tensorized Linear Layers}
The tensor contraction inside a TT-linear layer is highly sequential. Fortunately, our BTT-linear layer eliminates the data dependency in the first $d-1$ contraction steps, which brings parallelism inside the tensor linear layers and reduces the total length of contraction from $2d$ to $d+1$. 
% We denote this parallel BTT contraction as PBTT Table \ref{table:complex}. In this parallel setting, PBTT becomes more computing efficient than the right-to-left TT contraction when $n < 2K$.
However, directly applying intra-layer parallelism without considering inter-layer dataflow will cause inefficient resource usage. 
We propose task rescheduling and tensor fusion to tackle this problem.

\subsubsection{Task Rescheduling for Attention Layers} 
\begin{figure}[t]
    \centering
    % \includesvg[width=.5\textwidth]{fig_up/PBTT-FP.svg}
    % \includegraphics[width=.5\textwidth]{fig_up/PBTT-FP.png}
    \includegraphics[width=.55\textwidth]{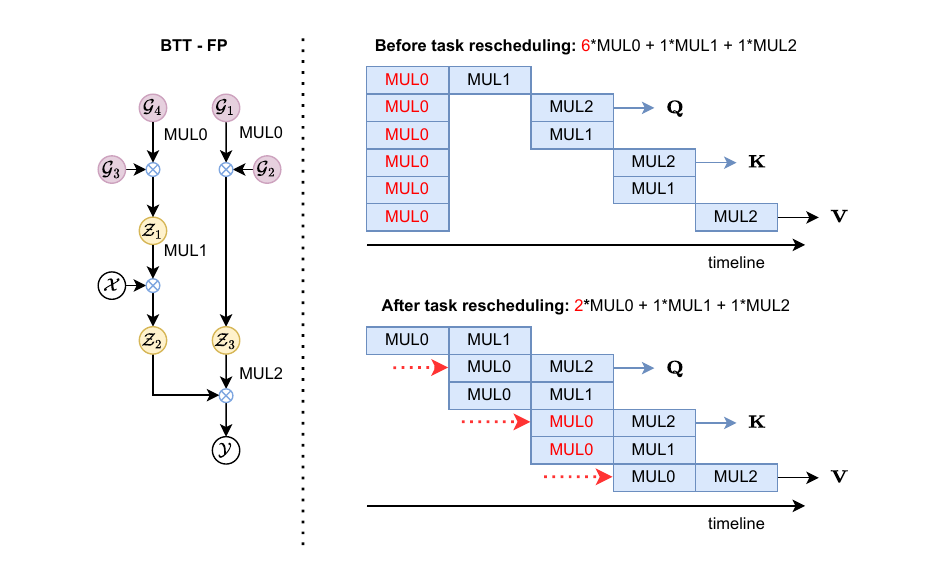}
    \caption{\textcolor{blue}{Schedule visualization of the $\mat{Q}$, $\mat{K}$, and $\mat{V}$ linear layers in BTT-format forward propagation before and after task rescheduling.
    Left: Computational graph for a single linear layer in BTT forward propagation.
    Right: Timeline-based schedule indicating the number of required compute kernels, with key resource usage (e.g., MUL0 units) highlighted in red.}}
    \label{fig:PBTT-FP}
\end{figure}

Fig. \ref{fig:PBTT-FP} compares the rescheduled parallel BTT and the default scheme for forward propagation of an attention block.
Here, we define the contraction kernels as follows:
\begin{itemize}[leftmargin=*]
    \item {\bf Kernel MUL0}: This kernel denotes the contraction between $\ten{G}_4$ and $\ten{G}_3$ that produces $\ten{Z}_1$, or the contraction between $\ten{G}_1$ and $\ten{G}_2$ which produces $\ten{Z}_3$. 
    
\item {\bf Kernel MUL1}: this kernel denotes the contraction between the input tensor $\ten{X}$ and the intermediate tensor $\ten{Z}_1$ that produces $\ten{Z}_2$.

\item {\bf Kernel MUL2}: this kernel denotes the contraction between intermediate tensors $\ten{Z}_2$ and $\ten{Z}_3$, producing the output of a linear layer $\ten{Y}$. In an attention block, $\ten{Y}$ can be the high-order tensorization of $\mat{Q}$, $\mat{K}$ or $\mat{V}$ features. 

\end{itemize}

% Meanwhile, the training process is also sequential across different TT-linear layer. Therefore, the FPGA resources will not be utilized efficiently if only intra-layer parallelism is considered. 
% To tackle this problem, we further explore the inter-layer parallelism by task rescheduling. \zz{Here do you mean the computing of $\mat{Q}$, $\mat{K}$ and $\mat{V}$ only when you say ``inter-layer" parallelism?}
% As shown in Fig.~\ref{fig:PBTT-FP}, we use the same index for the multipliers with similar computing scheme, which could share the same computing kernel in different times steps.

% To enable the parallelism of our bi-directional TT contraction, multiple PE arrays are created for each training stage for task-level parallelism. 
% In the first $d-1$ contraction steps of a forward pass, we contract the left and right $d$ TT cores in parallel (as shown in left and right columns in BTT graph separately) to get the first $2d-2$ intermediate results and store them in on-chip memory. In backward propagation, we compute the gradients of the left and right $d$ TT cores in parallel. After that, we update these TT cores in parallel immediately to avoid the overhead of storing the TT gradients.

A naive parallelization (Fig. \ref{fig:PBTT-FP} top right) of the $\mat{Q}$, $\mat{K}$ and $\mat{V}$ computation can lead to resource overhead on FPGA.
It runs all MUL0 kernels in the linear layers to compute $\mat{Q}$, $\mat{K}$ and $\mat{V}$ simultaneously since they are independent. As a result, six MUL0 kernels are needed in total. 
We use task rescheduling to reduce the hardware cost while achieving the same speedup on FPGA.  
% parallelism could also be used across different layers of a transformer. 
% For instance, the first $d-1$ contraction steps in the forward passes of query, key, and value in an attention layer could run in parallel and achieve $3\times$ speedup. 
% However, it is inefficient to create multiple kernels to parallelize one task since it will introduce many idle computing kernels during the training, which results in additional energy costs and low power efficiency. A better task-level parallelism design is required for better power-latency trade-off. 
The optimized tasks scheduling moves non-urgent MUL0 to later time steps and run them with other multipliers in parallel without increasing the total latency. This requires only $2$ (rather than $6$) reusable computing kernels for MUL0, leading to better resource utilization.

\subsubsection{Fused Parallel BTT} 
\begin{figure}[t]
    \hfill
    \includegraphics[width=.55\textwidth]{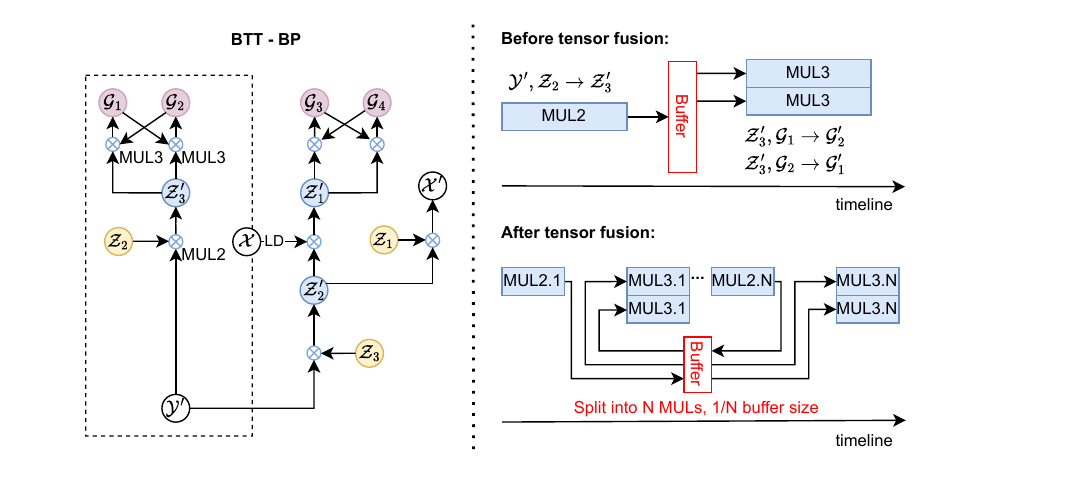}
    \caption{\textcolor{blue}{Schedule viewer of BTT-format back propagation before and after tensor fusion. Left: Computational graph of the BTT-linear layer during backpropagation.
    Right: Timeline-based execution schedule illustrating how tensor fusion enables buffer reuse and reduces buffer size, with key improvements highlighted in red.}}
    \label{fig:PBTT-BP}
\end{figure}

The intermediate results inside a tensor linear layer consume much more memory than the compressed model parameters. 
% Intermediate results is dominant inside TT-linear layer: Fine-grained dataflow to eliminate the memory overhead of intermediate results inside TT-linear layer. Need to re-design Fig. 9 to make it as simple as possible.
% Allocating BRAM blocks to N TT cores separately is not efficient. Solution: assign all TT cores to m=N/K groups. Assign BRAM blocks to each group. Just briefly mention the reshaping & partitioning. 
Therefore, we propose a fused parallel BTT dataflow to eliminate the memory overhead caused by the intermediate results, as shown in Fig. \ref{fig:PBTT-BP}. 
Here we define the contraction kernels as follows:
\begin{itemize}
    \item \textbf{Kernel MUL2:} this kernel contracts between the output gradient $\ten{Y}'$ and the intermediate result $\ten{Z}_2$, resulting in $\ten{Z}'_3$.
    \item \textbf{Kernel MUL3:} this kernel contracts the intermediate result $\ten{Z}'_3$ and the tensor factor ($\ten{G}_1$ or $\ten{G}_2$), produces $\ten{G}'_2$ or $\ten{G}'_1$, respectively, and then performs a parameter update.
\end{itemize}

We mainly apply operation fusion in these two contraction steps. As denoted in the original scheme (Fig. \ref{fig:PBTT-BP}. top), all multiplications for a tensor contraction are completed before the next contraction step. In this way, all intermediate results between contraction steps are required to be stored, which creates a large buffer with high memory consumption.
To relieve memory usage, we divide each contraction operation into multiple fine-grained contractions as follows:
\begin{itemize}
    \item \textbf{Fine-Grained Contraction for MUL2:} we perform contraction between sub-tensor $\ten{Y}'[:,i_1,i_2]$ with $\ten{Z}_2$ and produce subtensor $ \ten{Z}_3'[i_1,i_2,:]$.
    \item \textbf{Fine-Grained Contraction for MUL3:} we perform the contraction between a subtensor $\ten{Z}_3'[i_1,i_2,:]$ with another sub-tensor $\ten{G}_1[i_1,:]$ (or $\ten{G}_2[:,i_2,:]$), and produce a result sub-tensor $\ten{G}_2'[:,i_2,:]$ (or $\ten{G}_1'[i_1,:]$).
\end{itemize}

Here $i_1\in \{1,...,n_1\}, i_2\in \{1,...,n_2\}$, and we need to repeat the fine-grained multiplication for $n_1 n_2$ times.
% \zz{what do you mean by ``fine-grained" contractions?}\jiayi{fixed}.\zz{I still don't understand what specific operation you mean. Do you mean the multiplication of one slice inside one tensor contraction?} 
When one fine-grained contraction (Fig. \ref{fig:PBTT-BP} bottom) creates a small intermediate result, the next fine-grained contraction step uses it immediately, and thus only a small buffer with size $O(r)$ is required for all $n_1 n_2$ fine-grained contraction steps. The memory cost of intermediate contraction results is completely eliminated in the parameter gradient computation process.

\begin{figure}[t]
     \centering
     \includegraphics[width=.5\textwidth]{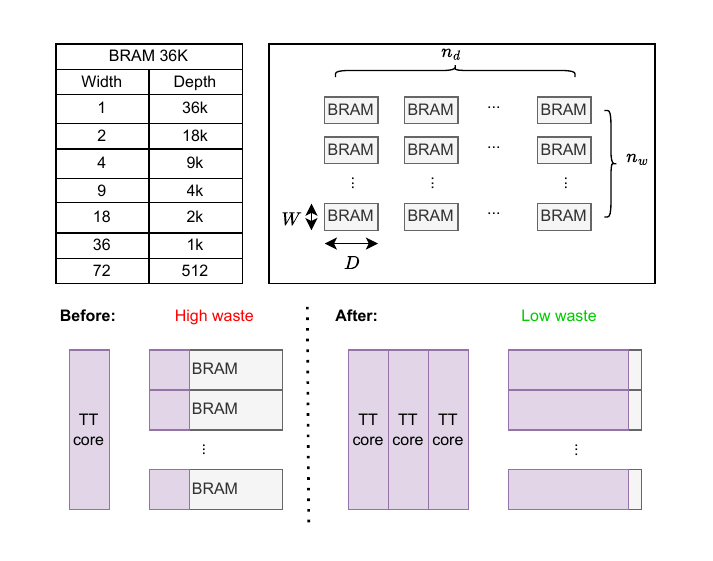}
    \caption{Configurations of BRAM 36K. Number of BRAM of one array. BRAM usage efficiency before and after tensor grouping.}
    % \zz{Better to show the high-level idea of the two methods only in the figure. The equations can be moved to the body texts.}}
    \label{fig:bram}
\end{figure}

% CU design for on-chip meomry saving
\subsection{Memory Management on FPGA}
\label{sec:memory}
% While the tensor-compressed training achieves an order-of-magnitude compression ratio in the model parameters, the on-chip memory resources should still be carefully used due to the additional intermediate results required to be stored and reused in training. 
Although tensor-compressed training allows storing all model parameters in the BRAM, creating separate memory blocks for multiple small-size TT cores can result in dramatic memory waste because of the fixed BRAM block size on FPGA.
% To address this issue, we reduce on-chip BRAM cost using a TT core grouping method.
We propose a tensor core grouping method to improve BRAM utilization efficiency.

Each BRAM block in the FPGA has a fixed storage size $C$, and can be configured with various widths ($W$) and depths ($D$) with $C=W\times D$.
For example, $C=36,864$ bits for most state-of-the-art AMD FPGA. A BRAM block can be configured as $W=1\sim 72$ as shown in Fig.~\ref{fig:bram} (top left). 
In order to store a large data array, $n_w\times n_d$ BRAM blocks are concatenated in the dimension of ``depth" or ``width", as shown in Fig.~\ref{fig:bram} (top right).
Here, $n_w$ is calculated by the width of the data array divided by $W$, $n_d$ is calculated by the depth of the data array divided by $D$. Both should be rounded up to their nearest integers, respectively. The data array width and depth are decided by the dimensions of the TT core (or grouped TT cores) after it is reshaped into a 2D structure.
% \zz{What is array width and array depth? We need to define them clearly.}

% \zz{original TT core or the intermediate tensor results?}
To efficiently store the TT cores, we first consider the influence of parallel computing inside each tensor contraction operation. In our design, we implement parallelism over the rank index across all tensor contractions involving TT cores, which requires $r$ parallel read operations from each TT core.
However, since each BRAM block has only two read/write ports, storing an entire TT core in a single BRAM block cannot support parallel access. 
High-Level Synthesis (HLS) offers two solutions to address this issue. The first approach, known as array partitioning, could partition a large data array into multiple small arrays and map them to separate BRAM blocks. Therefore, we use $r$ BRAM blocks per TT core to enable parallel data loading.
% \zz{Better to move the GEMV computing to here, and describe it exactly with one equation.}
The second approach, known as array reshaping,  concatenates multiple elements of the array by increasing the bit-width in order to support parallel data processing. 
This method is more memory efficient because it only consumes one BRAM block when $B_wr < {\rm max}(W)$, or $\lceil\frac{B_wr}{{\rm max}(W)}\rceil$ blocks otherwise. In this way, since we use floating-point 32, where $B_w = 32$ ($< {\rm max}(W) = 72$), the required number of BRAM blocks is always smaller than $r$.
% The two array reorganization methods are illustrated in Fig.~\ref{fig:bram}.

\begin{figure}[t]
     \centering
     \includegraphics[width=.45\textwidth]{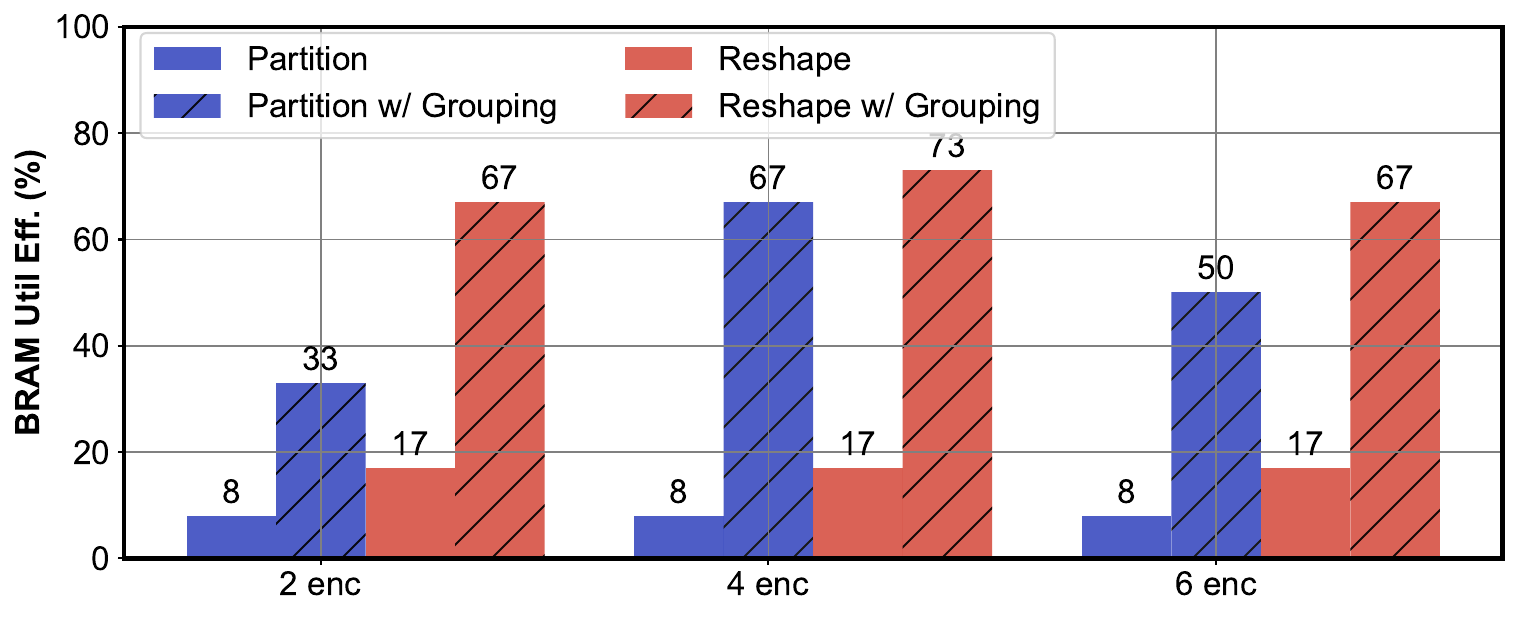}
    \caption{Comparison of the BRAM utilization efficiency using different memory management strategies in different model sizes.}
    %\zz{Reorganize this figure. Specifically, have four groups (based on the encoder numbers and FP formats), compare four management methods inside each group. } 
    \label{fig:bram_eff}
\end{figure}

Assuming that the tensor-compressed transformer involves $N$ TT cores having the same size for simplicity, the total number of BRAM blocks can be roughly estimated as $N n_w\times n_d$. 
% \zz{Do you assume that all TT cores have the same size?} %which could be calculated by $2d\times t \times l$.
% \zz{The figure should not show the equation. It should only show the high-level idea of the two methods. Then you should use one paragraph with key equation to analyze the memory complexity of the two methods, and how much saving is achieved compared with standard methods. }
Let $\lceil x \rceil$ denote rounding up $x$ to its nearest integer, then $n_w$ and $n_d$ can be decided as
\begin{align}
    &\text{Array\ partitioning: } n_w=r\lceil{\frac{B_w}{W}}\rceil, n_d = \lceil{\frac{nr}{D}}\rceil, \label{eqa:0}\\
    &\text{Array\ reshaping: } n_w=\lceil{\frac{B_wr}{W}}\rceil, n_d = \lceil{\frac{nr}{D}}\rceil. \label{eqa:1}
\end{align}
The above analysis indicates that $n_d$ will always be $1$ since generally $nr << D$, leading to a huge waste of BRAM usage. 

To improve BRAM efficiency, we increase the depth of the array by grouping some TT cores. Specifically, we concatenate $K$ TT-cores without data dependency into a single array along the depth size, then assign one BRAM block to the whole group. 
\textcolor{blue}{In our implementation, TT cores are concatenated across encoder layers and contraction directions. Given a total of $L$ encoder layers and $2d$ tensor factors, we set $K=(d-1)L$.}
In this case, the total number of BRAM blocks required is $\frac{N}{K}n_w\times n_d$, with $n_w$ and $n_d$ decided below:
\begin{align}
    &\text{Array\ partitioning: } n_w=r\lceil{\frac{B_w}{W}}\rceil, n_d = \lceil{\frac{Knr}{D}}\rceil, \label{eqa:2}\\
    &\text{Array\ reshaping: } n_w=\lceil{\frac{B_wr}{W}}\rceil, n_d = \lceil{\frac{Knr}{D}}\rceil. \label{eqa:3}
\end{align}
\textcolor{blue}{Here, scaling up small $\frac{nr}{D}$ using $K$ brings $\frac{Knr}{D}$ closer to its rounded-up integer}, and thus we can greatly improve the BRAM utilization efficiency, as shown in Fig.~\ref{fig:bram} (bottom). 

% Different array reorganization methods, BRAM configuration and choices of N could bring huge influence in the BRAM utilization efficiency. 

% \zz{Need to be more rigorously defined. What is the specific form of $F$? Can we optimize $W,D$ and $K$ simultaneously? What optimizer did you use to solve this problem?} \jiayi{K is hyper-para, and we just traversing all possible $\theta$ to find the min F, which is described below.}
The BRAM management problem is formulated as
\begin{align}
    \underset{\theta}{\min}\ F(\theta,\beta) = Nn_wn_d, \theta = \{W,D\}, \beta = \{N,K,n,r,B_w\}. \nonumber
    %\label{eqa:bram-eqa}
\end{align}
With a tensorization setting $\beta$, we can find $W$ and $D$ to maximize the BRAM utilization according to Eq. \eqref{eqa:0} -- \eqref{eqa:3}.

% \zz{This paragraph should appear early. Then you talk about each stage of your 3-stage optimization.}
% To summarize, we propose a three-stage optimization technique to optimize the on-chip memory management, which firstly chooses partition or reshape array reorganization method, and then choose the best BRAM configurations, and finally we introduce another hyper-parameter K to increase utilization efficiency in BRAM depth. 

To show the effectiveness of our method, we define the efficiency of BRAM utilization as $\eta = \frac{N_{\rm min}}{N_{\rm total}}$, where $N_{\rm min}$ is the ideal total memory usage for all tensor factors without considering the capacity of each BRAM block, and $N_{\rm total}$ is obtained using our BRAM management scheme. 
Fig.~\ref{fig:bram_eff} shows that our BRAM management method can achieve a $3.9\times$ to $8.4\times$ higher utilization efficiency than the default under various partitioning strategies and model volume.

\begin{table}[t]
\caption{Configurations of each layer in uncompressed and tensorized Transformer.}
\resizebox{.48\textwidth}{!}{
\begin{tabular}{|c|c|c|c|c|}
\hline
\multicolumn{1}{|l|}{} & Format & Matrix shape & Tensor shape          & Rank \\ \hline
Embedding              & TTM    & $(1000, 768)$  & $((10,10,10), (12,8,8))$ & $30$   \\ \hline
Attention              & TT     & $(768, 768)$   & $(12,8,8,8,8,12)$       & $12$   \\ \hline
Feed-forward           & TT     & $(768, 768)$  & $(12,8,8,8,8,12)$     & $12$   \\ \hline
Classification         & TT     & $(768, 768)$   & $(12,8,8,8,8,12)$       & $12$   \\ \hline
\end{tabular}
}
\label{table:setting}
\end{table}

% \begin{table}[t]
% \caption{Performance of Tensor-compressed versus standard matrix-format Transformer trianing on ATIS dataset.}
% \centering
% \begin{tabular}{|c|c|c|c|}
% \hline
% & intent accuracy & slot accuray & \multicolumn{1}{l|}{size (MB)} \\ \hline
% standard (L2) & $95.2\%$   & $97.0\%$ & $36.8\ (1\ \times)$                                         \\ \hline
% tensor-compressed (L2) & $95.0\%$   & $96.2\%$ & $1.2\ (30.7\ \times)$                                \\ \hline
% standard (L6) & $93.6\%$   & $96.2\%$ & $93.6\ (1\ \times)$                                                  \\ \hline
% tensor-compressed (L6) & $95.0\%$   & $96.4\%$ & $1.8\ (52.0\ \times)$                                                              \\ \hline
% \end{tabular}
% \label{table:acc}
% \end{table}

% \begin{table}[t]
% \caption{Performance of Tensor-compressed versus standard matrix-format Transformer trianing on ATIS dataset.}
% \centering
% \begin{tabular}{|c|c|c|c|c|}
% \hline
% & Model & intent accuracy & slot accuray & \multicolumn{1}{l|}{size (MB)} \\ \hline
% 2-ENC & matrix & $95.2\%$   & $97.0\%$ & $18.4\ (1\ \times)$                                         \\ \hline
% tensor-compressed & $95.0\%$   & $96.2\%$ & $0.6\ (30.7\ \times)$                                \\ \hline
% matrix & $93.6\%$   & $96.2\%$ & $46.8\ (1\ \times)$                                                  \\ \hline
% tensor-compressed & $95.0\%$   & $96.4\%$ & $0.9\ (52.0\ \times)$                                                              \\ \hline
% \end{tabular}
% \label{table:acc}
% \end{table}

\begin{table}[]
\centering
\caption{Performance of Tensor-compressed versus standard matrix-format Transformer training on ATIS dataset.}
\begin{tabular}{|c|c|c|c|}
\hline
Model        & Intent \textcolor{blue}{Test} Acc & Slot \textcolor{blue}{Test} Acc & Size (MB)             \\ \hline
2-ENC matrix & $96.2\%$   & $\mathbf{97.3\%}$ & $36.7$                \\ \hline
\textbf{2-ENC tensor} & $\mathbf{97.0\%}$   & $97.2\%$ & $\mathbf{1.2 \ (30.5 \times)}$ \\ \hline
4-ENC matrix & $95.7\%$   & $\mathbf{97.4\%}$ & $65.1$                \\ \hline
\textbf{4-ENC tensor} & $\mathbf{96.6\%}$   & $\mathbf{97.4\%}$ & $\mathbf{1.5 \ (43.4 \times)}$ \\ \hline
6-ENC matrix & $96.4\%$   & $\mathbf{97.5\%}$ & $93.5$                \\ \hline
\textbf{6-ENC tensor} & $\mathbf{97.0\%}$   & $97.2\%$ & $\mathbf{1.8 \ (52.0 \times)}$ \\ \hline
\end{tabular}
\label{table:acc}
\end{table}

\section{Experiments}
\subsection{Experimental Setup}
We implement our tensor-compressed transformer training accelerator with high-level synthesis (HLS) using C++, and perform synthesis, placement, and routing using Vitis HLS 2023.2 on an AMD Alveo U50 FPGA board. The programmable logic fabric has $872$k LUTs, $5952$ DSPs, $5.9$-MB BRAMs, and $22.5$-MB URAMs. All model parameters, activations, and gradients during training are implemented in the floating-point $32$-bit format with $100$-MHz frequency.
% To support more memory and computational intensive models and tasks, we could also customize the data type to floating-point 16 with 125MHz for lower memory and computing burden. 

\subsection{\textcolor{blue}{Algorithmic Evaluation}}
We evaluated the FPGA accelerator by training a transformer on the ATIS dataset \cite{hemphill1990atis}. This is a data set widely used in natural language understanding (NLU), including audio recordings of customers booking flight tickets.
The input sequence length is set to $32$, but the accelerator is also re-configurable to support larger sequence length and for other datasets. For functionality evaluation, we use the stochastic gradient descent (SGD) optimizer with a learning rate $4\times 10^{-3}$ and batch size of $1$. The hyper-parameters of uncompressed training and our tensor-compressed training are listed in Table \ref{table:setting}. 
Table \ref{table:acc} shows that the tensor-compressed method could achieve the same accuracy with $30.5\times$ to $52.0 \times$ compression ratios for transformers with $2$ to $6$ encoding blocks.
%To evaluate the functionality of our PBTT Transformer training accelerator, we also construct the whole training framework on PyTorch and compare the loss and accuracy curve of python and c simulation. 

% Since there is no existing work for Transformer model training on FPGA, we compare the throughput and energy-efficiency with some state-of-the-art CNN and DNN training accelerators conducting floating-point training on FPGA.

\begin{figure}[t]
\centering
\includegraphics[width=.5\textwidth]{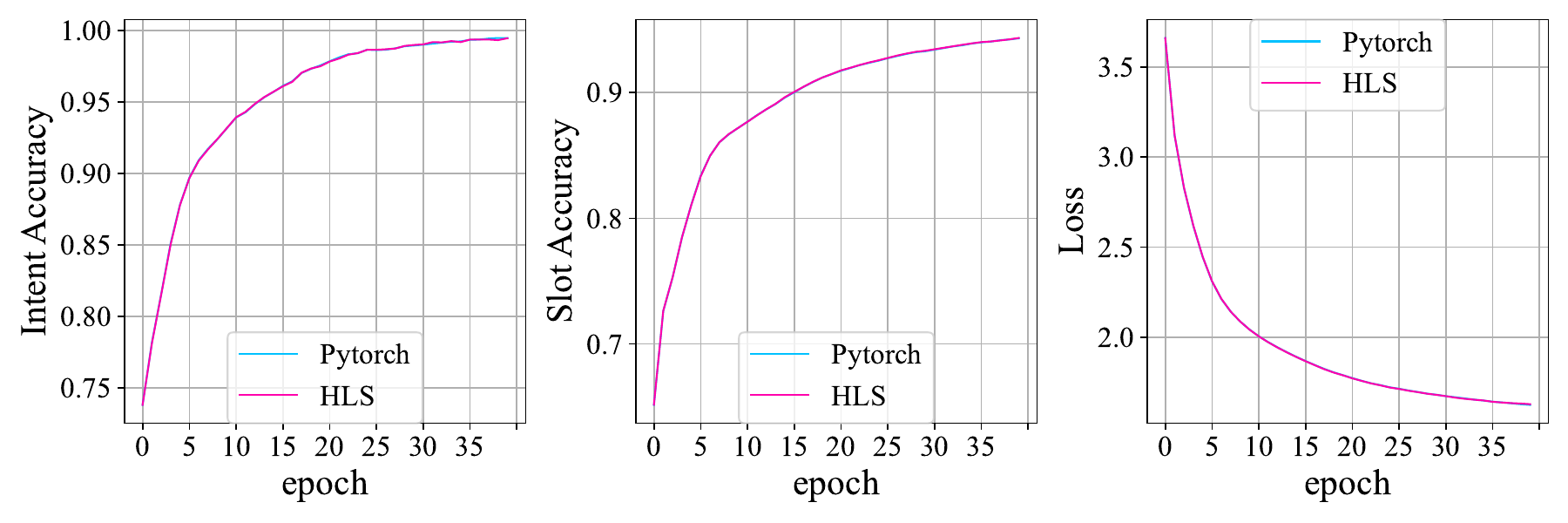}
\caption{Training accuracy and loss of the accelerator (2-ENC) compared with PyTorch in ATIS dataset through the whole training process (40 epochs).}
% \zz{(1) Change the legends to ``Pytorch" and ``HLS". (2) change the y-label to ``classification" and "slot" in the first two sub-figures. }}
% right three figures zoom in the training curves of the last 20 epochs. 
%\zz{No need the results of the last 20 epochs. Organize the results of the whole training process as a figure with three sub-figures in the horizontal direction. }}
\label{fig:corr}
\end{figure}

\begin{figure}[t]
     \centering
     \includegraphics[width=.5\textwidth]{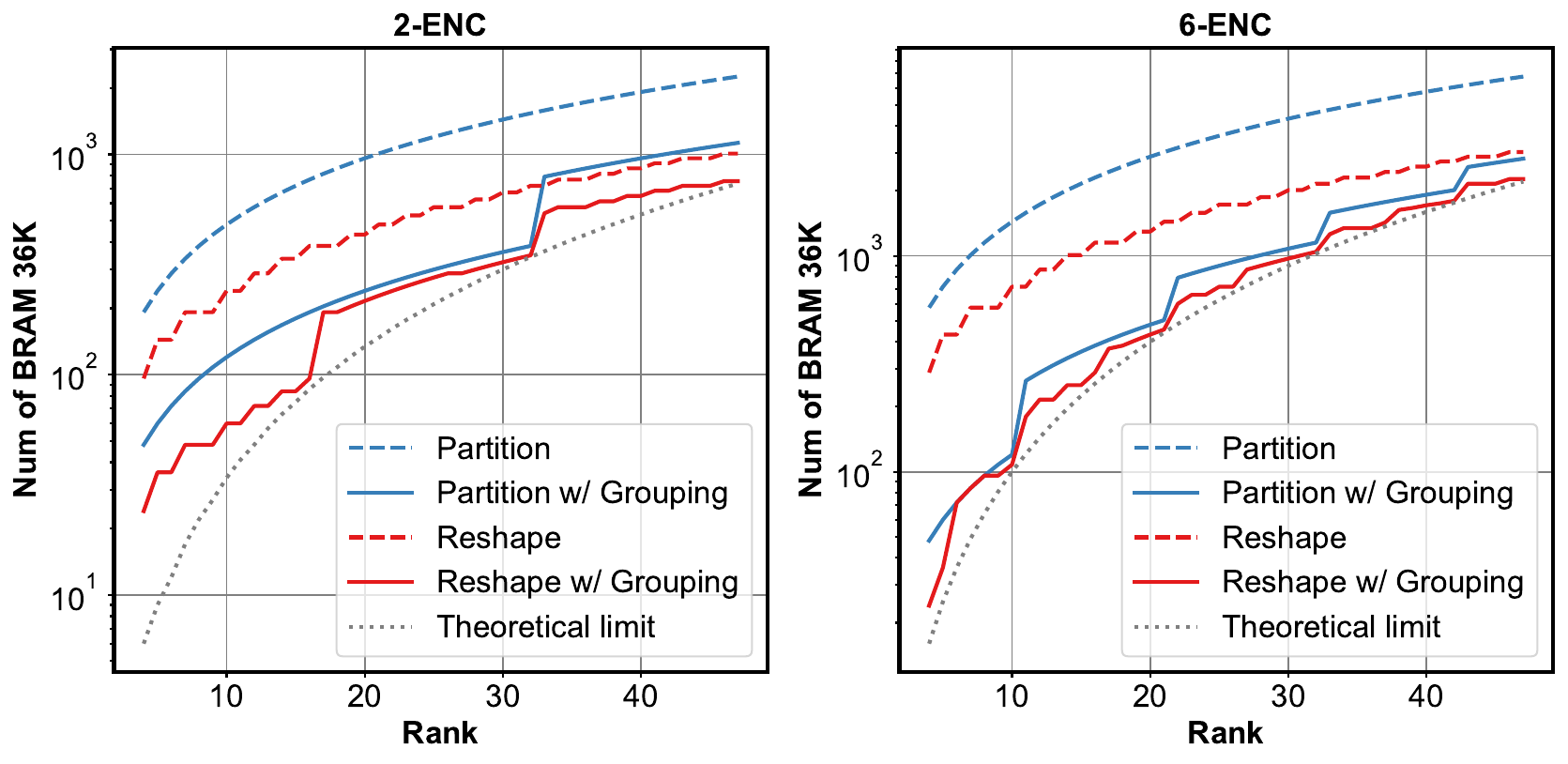}
    \caption{BRAM usage for all tensor factors using different memory management strategy versus rank.}
    \label{fig:bram_comp}
\end{figure}

\subsection{FPGA Accelerator Evaluation}
\subsubsection{Functionality Evaluation}
To validate the functionality of our transformer training accelerator with parallel BTT implementation, we compare the training loss and accuracy curves of our high-level synthesis (HLS) FPGA implementation with PyTorch training on a GPU, as illustrated in Fig.~\ref{fig:corr}. The ATIS dataset is a multi-task dataset for both intent classification and slot filling. 
Intent classification aims to determine the overarching goal or purpose of the user’s query and is treated as a text classification problem.
Slot filling identifies key entities or attributes (slots) within a user’s utterance and classifies them into predefined categories.
The accuracy and loss curves for both tasks demonstrate that the results from our HLS FPGA implementation closely match those from PyTorch training, confirming the consistency and reliability of our accelerator design.
% Additionally, the right three figures zoom in the curve in last 20 epochs and we could observe a slight difference between SW and HW implementations. The slight difference is caused by machine error introduced by different computing order and will not influence the whole training process. 

\begin{figure}[t]
     \centering
     \includegraphics[width=.5\textwidth]{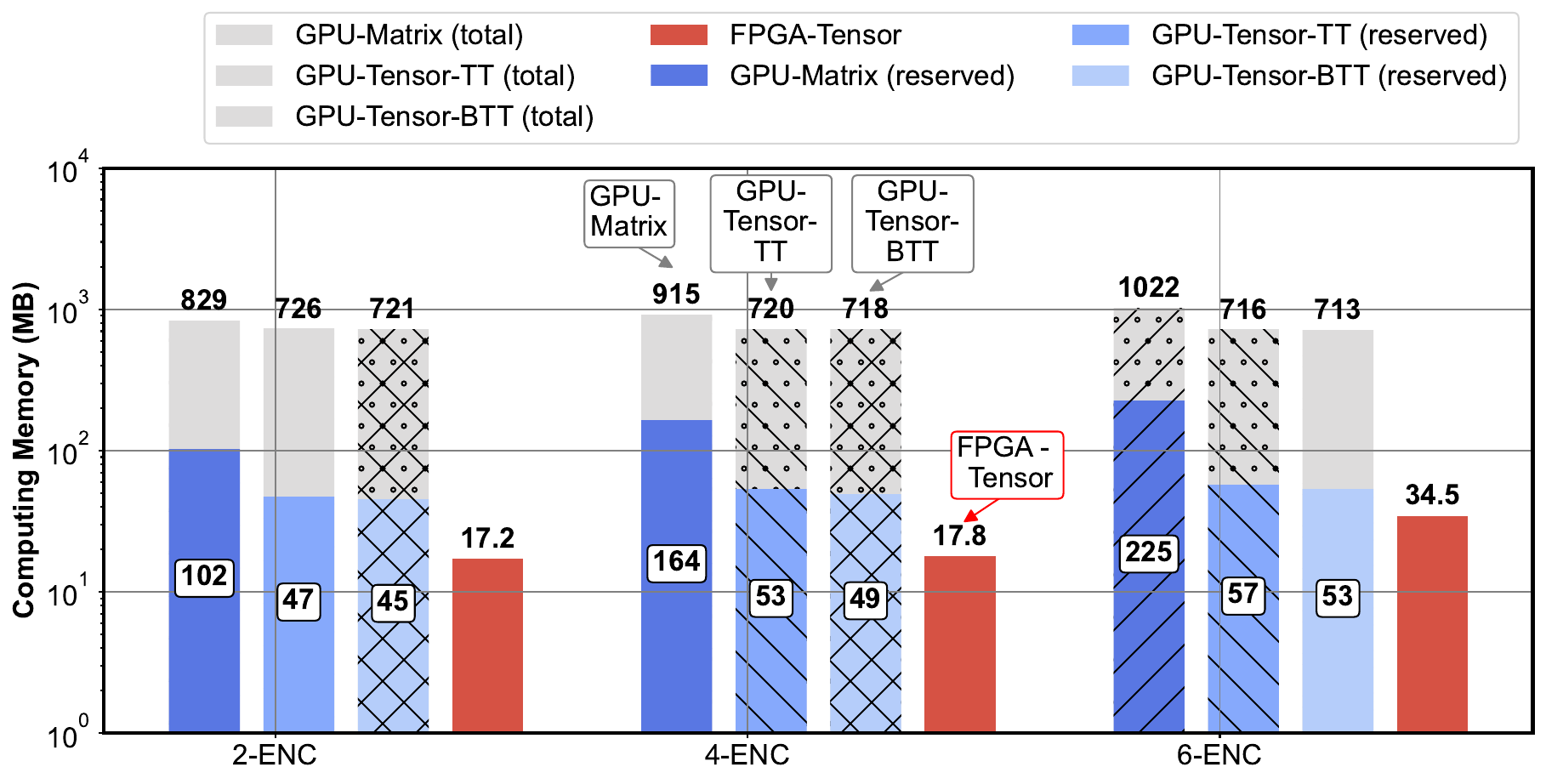}
    \caption{\textcolor{blue}{Comparison of the computing memory costs between GPU  (with or without additional overhead) and FPGA in different model size.}
    % Here L2 and L6 mean a transformer with $2$ or $6$ encoding blocks; S32 and S64 mean the sequence length is 32 or 64.
    }
   % \end{center}
    \label{fig:matrix}
\end{figure}

\subsubsection{Effects of BRAM Optimization}
Fig.~\ref{fig:bram_comp} illustrates the number of BRAMs required to store all TT cores using different allocation methods in Section \ref{sec:memory}. 
% in various model settings, including fp32 or fp16 data formats and 2 or 6 encoder layers. 
It is shown that our proposed BRAM management strategy always consumes the least BRAM resources and even reaches the theoretical limit in some specific rank settings. Our method could bring the actual cost of BRAM closer to the ideal usage, because the TT-core grouping could reduce the waste of depth per BRAM block. 
Combining this BRAM optimization with tensor compression, the memory burden is further relieved, allowing storing all compressed model parameters in on-chip memory.

\begin{table*}[t]
\caption{Resource utilization and power consumption with various model structure settings and input workload.}
\centering
\begin{tabular}{|c|c|c|c|c|c|clcc|}
\hline
\multirow{2}{*}{Model Type} & \multirow{2}{*}{DSP} & \multirow{2}{*}{LUT} & \multirow{2}{*}{FF} & \multirow{2}{*}{BRAM} & \multirow{2}{*}{URAM} & \multicolumn{4}{c|}{Power (W)}                                     \\ \cline{7-10} 
                            &                      &                      &                     &                       &                       & \multicolumn{2}{c|}{Dynamic} & \multicolumn{1}{c|}{Static} & Total \\ \hline
2-ENC                       & $2396 \ (40\%)$      & $565\text{k} \ (65\%)$ & $475\text{k} \ (27\%)$ & $1216 \ (90\%)$         & $114 \ (18\%)$          & \multicolumn{2}{c|}{$20.68$} & \multicolumn{1}{c|}{$6.00$}    & $26.68$ \\ \hline
4-ENC                       & $2396 \ (40\%)$      & $572\text{k} \ (66\%)$ & $485\text{k} \ (28\%)$ & $1163 \ (86\%)$         & $128 \ (20\%)$          & \multicolumn{2}{c|}{$20.85$} & \multicolumn{1}{c|}{$5.97$} & $26.82$ \\ \hline
6-ENC                       & $2396 \ (40\%)$      & $579\text{k} \ (67\%)$ & $499\text{k} \ (29\%)$ & $1089 \ (81\%)$         & $374 \ (58\%)$          & \multicolumn{2}{c|}{$20.97$} & \multicolumn{1}{c|}{$6.09$} & $27.06$ \\ \hline
Available                   & $5952$               & $872\text{k}$         & $1,743\text{k}$       & $1344$                  & $640$                   & \multicolumn{4}{c|}{$75$}                                         \\ \hline
\end{tabular}
\label{table:usage}
\end{table*}

\subsubsection{Overall Accelerator Performance}
Table \ref{table:usage} presents the hardware resource utilization of our FPGA training accelerator in various model sizes from RTL implementation reports. We evaluated the accelerator with a transformer with $2$ to $6$ encoder layers. 
The utilization of DSPs, LUTs, and FFs remains consistent across different model configurations, as the same computational kernels are employed regardless of the model settings. However, BRAM usage decreases while URAM usage increases with a greater number of layers. This behavior can be attributed to the application of our grouping technique to intermediate arrays, which demands additional memory as the number of layers grows. Consequently, HLS automatically allocates these memory requirements to URAM to enhance the utilization efficiency of on-chip memory.

\subsection{Performance Comparison with GPU}
% \zz{Organize as a single subsection to hilight the comparison with GPU.}
We perform end-to-end performance evaluation of our FPGA accelerator (operating at 100 MHz) compared to the NVIDIA RTX 3090 GPU (acting at 1395 MHz). 
The evaluation includes metrics such as memory usage, latency, power consumption, and overall energy efficiency across both platforms. 
\textcolor{blue}{Specifically, we benchmark matrix-based, sequential TT-based, and bidirectional TT (BTT)-based training on the GPU, and BTT-based training on our FPGA accelerator.} 
For each setup, we report the total system-level power consumption, measured during full training execution.
%The latency per iteration is calculated by the total cycles of operations generated from HLS synthesis multiplied by the frequency, and we could get the latency per epoch by multiplying the number of input sample batch. 
%The GPU power consumption is from NVIDIA-SMI instruction. The FPGA power consumption is from the post-implementation reports.

\subsubsection{Memory Cost} Fig.~\ref{fig:matrix} shows the memory cost of the RTX 3090 GPU and our FPGA accelerator on the ATIS dataset with different model configurations. 
For GPU memory, we report two metrics: total memory (extracted from the ${\rm nvidia}$-${\rm smi}$ command) and reserved memory (extracted with the CUDA ${\rm memory}\_{\rm reserved}$ functions). 
The former measures the memory cost in the entire training process, \textcolor{blue}{encompassing model allocations as well as framework-level overheads. In contrast, the reserved memory reflects only the portion occupied by model parameters, activations, and optimizer states.}
The computing memory of FPGA is the on-chip memory consumption collected from the RTL implementation report.
\textcolor{blue}{Unlike GPUs, FPGAs allow for explicit allocation of BRAM and URAM, offering deterministic and fragmentation-free memory usage.}

Benefiting from the hardware and algorithm co-design for tensor-compressed training, our FPGA accelerator can perform on-device training with $20.7\times$ to $42.2\times$ lower memory cost compared to tensor-compressed training on GPU. 
\textcolor{blue}{When excluding the framework-level overhead on the GPU, our BTT algorithm achieves a $2.3\times$ to $4.2\times$ memory reduction on the GPU compared to a standard matrix-based Transformer. Furthermore, our fine-grained on-chip memory management offers an additional $1.5\times$ to $2.7\times$ memory reduction relative to the GPU’s reserved memory footprint.}

% \usepackage{multirow}
\begin{table*}[htbp]
\caption{\textcolor{blue}{Performance comparison between GPU and our FPGA accelerator in various model settings and contraction types.}}
\centering
\begin{tabular}{|c|c|c|c|cc|cc|}
\hline
\multirow{2}{*}{Model Settings} & \multirow{2}{*}{Platform-Type} & \multirow{2}{*}{Latency per epoch (sec)} & \multirow{2}{*}{Power (W)} & \multicolumn{2}{c|}{Computing Memory}      & \multicolumn{2}{c|}{Energy per epoch}      \\ \cline{5-8} 
                                &                                &                                                                                     &                            & \multicolumn{1}{c|}{(MB)} & \textbf{Ratio ($\times$) }  & \multicolumn{1}{c|}{(kJ)} & \textbf{Ratio ($\times$) } \\ \hline
\multirow{4}{*}{L2-S32-FP32}    & GPU-Matrix                     & 47                                                                                  & 150                        & \multicolumn{1}{c|}{829}  & \textbf{48.2}  & \multicolumn{1}{c|}{7.1}  & \textbf{1.38}  \\
                                & \textcolor{blue}{GPU-TT}                         & 144                                                                                 & 140                        & \multicolumn{1}{c|}{726}  & \textbf{42.2}  & \multicolumn{1}{c|}{20.2} & \textbf{3.96}  \\
                                & GPU-BTT                        & 129                                                                                 & 138                        & \multicolumn{1}{c|}{721}  & \textbf{41.9}  & \multicolumn{1}{c|}{17.8} & \textbf{3.49}  \\ \cline{2-8} 
                                & \textbf{FPGA-BTT (ours)}                       & 191                                                                                 & 26.68                      & \multicolumn{1}{c|}{17.2} & \textbf{1.0}     & \multicolumn{1}{c|}{5.1}  & \textbf{1.0}     \\ \hline
\multirow{4}{*}{L4-S32-FP32}    & GPU-Matrix                     & 77                                                                                  & 150                        & \multicolumn{1}{c|}{915}  & \textbf{51.4}  & \multicolumn{1}{c|}{11.6} & \textbf{1.29}  \\
                                & \textcolor{blue}{GPU-TT}                         & 243                                                                                 & 138                        & \multicolumn{1}{c|}{720}  & \textbf{40.4}  & \multicolumn{1}{c|}{33.5} & \textbf{3.73}  \\
                                & GPU-BTT                        & 222                                                                                 & 138                        & \multicolumn{1}{c|}{718}  & \textbf{40.3}  & \multicolumn{1}{c|}{30.6} & \textbf{3.41}  \\ \cline{2-8} 
                                & \textbf{FPGA-BTT (ours)}                       & 335                                                                                 & 26.82                           & \multicolumn{1}{c|}{17.8} & \textbf{1.0}     & \multicolumn{1}{c|}{9.0}  & \textbf{1.0}     \\ \hline
\multirow{4}{*}{L6-S32-FP32}    & GPU-Matrix                     & 108                                                                                 & 152                        & \multicolumn{1}{c|}{1022} & \textbf{29.6}  & \multicolumn{1}{c|}{16.4} & \textbf{1.26}  \\
                                & \textcolor{blue}{GPU-TT}                         & 347                                                                                 & 138                        & \multicolumn{1}{c|}{716}  & \textbf{20.8}  & \multicolumn{1}{c|}{47.9} & \textbf{3.67}  \\
                                & GPU-BTT                        & 324                                                                                 & 138                        & \multicolumn{1}{c|}{713}  & \textbf{20.7}  & \multicolumn{1}{c|}{44.7} & \textbf{3.43}  \\ \cline{2-8} 
                                & \textbf{FPGA-BTT (ours)}                       & 482                                                                                 & 27.06                      & \multicolumn{1}{c|}{34.5} & \textbf{1.0}     & \multicolumn{1}{c|}{13.0} & \textbf{1.0}     \\ \hline
\end{tabular}
\label{table:GPU}
\end{table*}

\subsubsection{Energy Cost} Table~\ref{table:GPU} further summarizes the latency per epoch, power consumption, and energy per training epoch of our FPGA accelerator collected from HLS synthesis and RTL implementation reports, compared to both uncompressed training and tensor-compressed training on the GPU. 
\textcolor{blue}{On GPU, BTT achieves lower energy consumption and reduced compute memory usage compared to sequential TT, but the improvements remain relatively modest.}
Due to the huge difference in clock frequency, our FPGA accelerator has a higher training latency. 
Nevertheless, it achieves \textcolor{blue}{over $3.6\times$ and $3.4\times$ lower energy consumption than TT and BTT training on GPU, respectively.}
We also compare against standard matrix-based GPU training, which benefits from optimized hardware support for matrix operations. While this matrix-format GPU training is more energy-efficient than tensor-compressed GPU training, our FPGA accelerator still achieves approximately $1.3\times$ lower energy consumption.
% , while using $29.6\times$ to $51.4\times$ less compute memory.

\subsubsection{Comparison with Previous Work} Some previous works \cite{liu2017fpga, venkataramanaiah2019automatic, venkataramanaiah2020fpga} achieved higher energy efficiency improvement compared with GPU. However, these works are all CNN training accelerators supporting small models, and their GPU baseline is less powerful than ours. 
For instance, Ref. \cite{liu2017fpga} using Xilinx ZU19EG FPGA achieved $8.5\times$ improvement in energy efficiency in LeNet-10 compared to the GTX 1080 Ti GPU. However, LeNet-10 has $129 \times$ to $328 \times$ fewer model parameters and requires $22\times$ to $61\times$ lower total computing costs than our transformer training. 
Similarly, the work in \cite{venkataramanaiah2020fpga} using Intel Stratix-10 MX FPGA achieved $4.5\times$ higher energy efficiency than Tesla V100 in training the ResNet-20 model. Our transformer model is $41\times$ to $106\times$ larger than ResNet-20,  and our computing cost is $6.9\times$ to $18.7\times$ higher. Compared with these results, our accelerator can achieve higher energy efficiency than the more powerful RTX 3090 GPU, and can train much larger models with limited memory.

% \input{limitation}

% \begin{algorithm}[H]
% \caption{Weighted Tanimoto ELM.}\label{alg:alg1}
% \begin{algorithmic}
% \STATE 
% \STATE {\textsc{TRAIN}}$(\mathbf{X} \mathbf{T})$
% \STATE \hspace{0.5cm}$ \textbf{select randomly } W \subset \mathbf{X}  $
% \STATE \hspace{0.5cm}$ N_\mathbf{t} \gets | \{ i : \mathbf{t}_i = \mathbf{t} \} | $ \textbf{ for } $ \mathbf{t}= -1,+1 $
% \STATE \hspace{0.5cm}$ B_i \gets \sqrt{ \textsc{max}(N_{-1},N_{+1}) / N_{\mathbf{t}_i} } $ \textbf{ for } $ i = 1,...,N $
% \STATE \hspace{0.5cm}$ \hat{\mathbf{H}} \gets  B \cdot (\mathbf{X}^T\textbf{W})/( \mathbb{1}\mathbf{X} + \mathbb{1}\textbf{W} - \mathbf{X}^T\textbf{W} ) $
% \STATE \hspace{0.5cm}$ \beta \gets \left ( I/C + \hat{\mathbf{H}}^T\hat{\mathbf{H}} \right )^{-1}(\hat{\mathbf{H}}^T B\cdot \mathbf{T})  $
% \STATE \hspace{0.5cm}\textbf{return}  $\textbf{W},  \beta $
% \STATE 
% \STATE {\textsc{PREDICT}}$(\mathbf{X} )$
% \STATE \hspace{0.5cm}$ \mathbf{H} \gets  (\mathbf{X}^T\textbf{W} )/( \mathbb{1}\mathbf{X}  + \mathbb{1}\textbf{W}- \mathbf{X}^T\textbf{W}  ) $
% \STATE \hspace{0.5cm}\textbf{return}  $\textsc{sign}( \mathbf{H} \beta )$
% \end{algorithmic}
% \label{alg1}
% \end{algorithm}

\section{Conclusion}
In this work, we have developed the first proof-of-concept end-to-end tensor-compressed transformer training accelerator on FPGA. We have identified a key limitation of the prior tensor network accelerators as the low-parallel computing flow and proposed a bidirectional flow with both lower computational costs and memory costs. 
We have further optimized dataflow and proposed a BRAM memory management method to achieve better latency, memory, and energy efficiency. The proposed tensorized transformer accelerator has achieved up to $4.0\times$ lower energy costs and $51\times$ lower computing memory costs than the Nvidia RTX 3090 GPU.
The model can scale up to $6$ encoder layers and has the potential to solve more complex training tasks on FPGA.

\section*{Acknowledgments}
This work is co-funded by Intel Strategic Research Sectors (SRS) - Systems Integration SRS \& Devices SRS.

% {\appendix[Complexity of TTM Contraction]}
% \label{sec:appendix}
% \input{appendix}

\bibliographystyle{IEEEtran}
\bibliography{ref}

\section{Biography Section}
% If you have an EPS/PDF photo (graphicx package needed), extra braces are needed around the contents of the optional argument to biography to prevent  the LaTeX parser from getting confused when it sees the complicated  $\backslash${\tt{includegraphics}} command within an optional argument. (You can create  your own custom macro containing the $\backslash${\tt{includegraphics}} command to make things  simpler here.)
 

\begin{IEEEbiography}[{\includegraphics[width=1in,height=1.25in,clip,keepaspectratio]{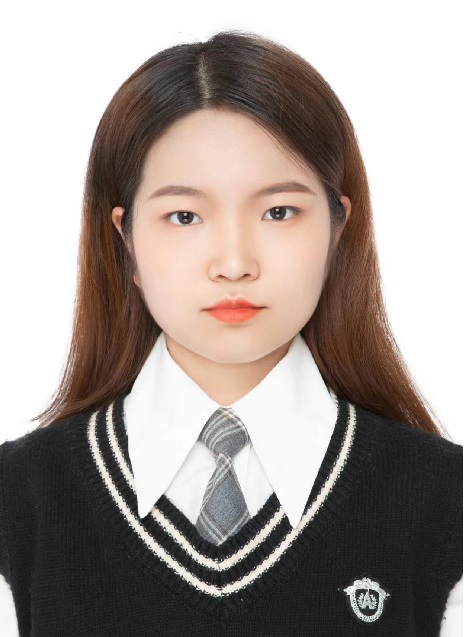}}]{Jiayi Tian}
Jiayi Tian received the B.Eng. degree in VLSI Design \& System Integration from Nanjing University, Nanjing, China, in 2023, and is currently a Ph.D. student in computer engineering in University of California, Santa Barbara, advised by Prof. Zheng Zhang. Her current research interests are algorithm \& hardware co-design for efficient large language model training and inference.
\end{IEEEbiography}

\begin{IEEEbiography}[{\includegraphics[width=1in,height=1.25in,clip,keepaspectratio]{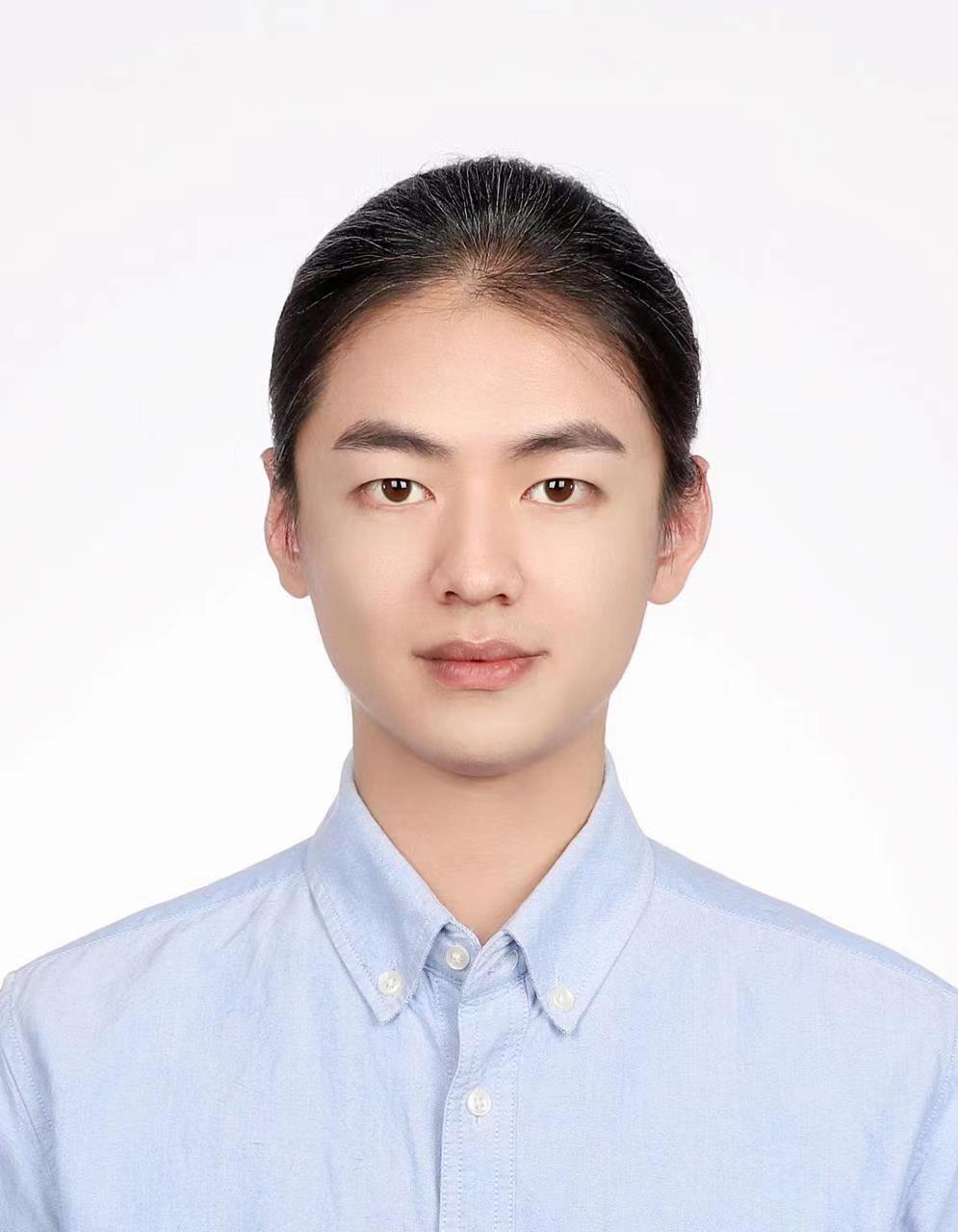}}]{Jinming Lu}
Jinming Lu received the B.S. degree in microelectronics from Nankai University, Tianjin, China,
in 2018, and the Ph.D. degree in information and communication engineering from Nanjing University, Nanjing, China, in 2023. Currently, he is a post-doc in University of California, Santa Barbara. His current research interests include automatic speech recognition and deep learning, especially its hardware acceleration and compression algorithms.
\end{IEEEbiography}

\begin{IEEEbiography}[{\includegraphics[width=1in,height=1.25in,clip,keepaspectratio]{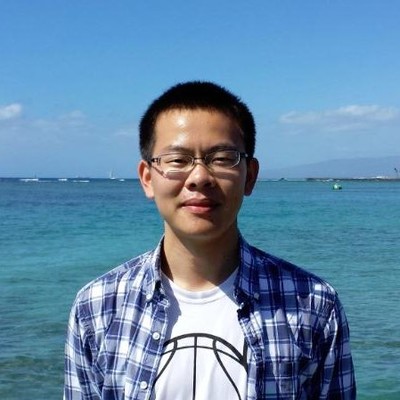}}]{Hai Li}
Hai Li received the B.S. degree in applied physics from the Huazhong University of Science and Technology, Wuhan, China, in 2011, and the M.S. and Ph.D. degrees in electrical and computer engineering from Carnegie Mellon University, Pittsburgh, PA, USA, in 2015 and 2016, respectively, where he developed foundational theorem of next generation storage system. He joined Intel Corporation in 2016, starting with specialized technologies program. He is currently a Senior Research Scientist with the Exploratory Integrated Circuits, Components Research, Intel Corporation, Hillsboro, OR, USA. He is working on emerging technologies in the joint area of novel device, circuit, and computing paradigm, while managing university collaboration. He serves as Co-Chair of Intel Emerging Technology Strategic Research Sector (SRS).
\end{IEEEbiography}

\begin{IEEEbiography}[{\includegraphics[width=1in,height=1.25in,clip,keepaspectratio]{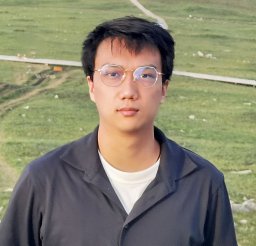}}]{Xiangwei Wang}
Xiangwei Wang received the B.Eng. degree in cyberspace security from Huazhong University of Science and Technology, Wuhan, China, in 2023, and currently is a Ph.D. student in the Department of Computer Science at North Carolina State University. He was an exchange student at the University of California, Santa Barbara (2022–2023), supervised by Prof. Zheng Zhang. 
His current research interests are programming languages, compiler and program optimization, and distributed systems for LLMs.
\end{IEEEbiography}

\begin{IEEEbiography}[{\includegraphics[width=1in,height=1.25in,clip,keepaspectratio]{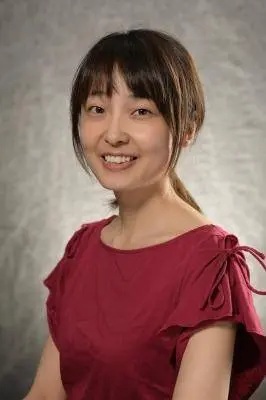}}]{Cong (Callie) Hao}
Cong (Callie) Hao (Associate Member, IEEE) received the Ph.D. degree from Waseda University, Tokyo, Japan, in 2017. She was a Postdoctoral Fellow with the Georgia Institute of Technology (Georgia Tech), Atlanta, GA, USA, from 2020 to 2021 and also worked as a Postdoctoral Researcher of ECE with the University of Illinois at Urbana-Champaign, Champaign, IL, USA, from 2018 to 2020. She is currently an Assistant Professor with the Department of Electrical and Computer Engineering, Georgia Tech. Her current research interests lie in the joint area of efficient hardware design and machine learning algorithms, reconfigurable and high-efficiency computing, and electronic design automation tools. Dr. Hao was a recipient of the NSF CAREER Award.
\end{IEEEbiography}

\begin{IEEEbiography}[{\includegraphics[width=1in,height=1.25in,clip,keepaspectratio]{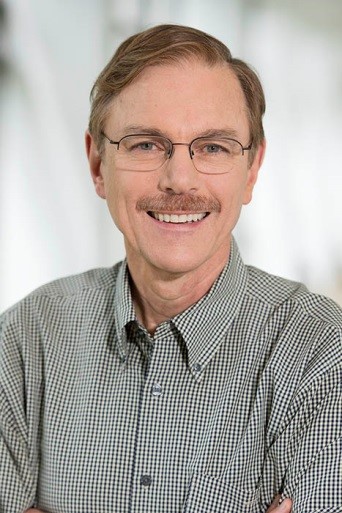}}]{Ian Young}
Ian. A. Young [M'788, SM96, F'99] is a senior fellow and director of Advanced Circuits and Technology Integration in the Technology and Manufacturing Group at Intel Corporation. 
He received his Bachelor's and Master's degrees in electrical engineering from the University of Melbourne, Australia, in 1972 and 1975. He received his Ph.D. in electrical engineering from the University of California, Berkeley in 1978. He joined Intel in 1983. 
Currently, he is responsible for defining future circuit directions with emerging novel devices and identifying leading options to manufacture energy efficient integrated circuits for computing in the beyond-CMOS era.
He has authored or coauthored more than 300 technical articles and holds over 250 U.S. patents.
Dr. Young has received three Intel Achievement Awards. He was a recipient of the 2009 International Solid-State Circuits Conference’s Jack Raper Award for outstanding technology directions paper and the 2018 IEEE Frederik Philips Award “for leadership in research and development on circuits and processes for the evolution of microprocessors.” 
% He served as the Technical
% Program Committee Chairperson of the 2005 International Solid-State Circuits
% Conference (ISSCC) and the Chairperson of the 1997 and 1998 Symposium on
% VLSI Circuits. He is a three time Guest Editor for Special Issues of the IEEE
% JOURNAL OF SOLID-STATE CIRCUITS and the Founder and the Inaugural
% Editor-in-Chief of the IEEE JOURNAL ON EXPLORATORY SOLID-STATE
% COMPUTATIONAL DEVICES AND CIRCUITS.
\end{IEEEbiography}

\begin{IEEEbiography}[{\includegraphics[width=1in,height=1.25in,clip,keepaspectratio]{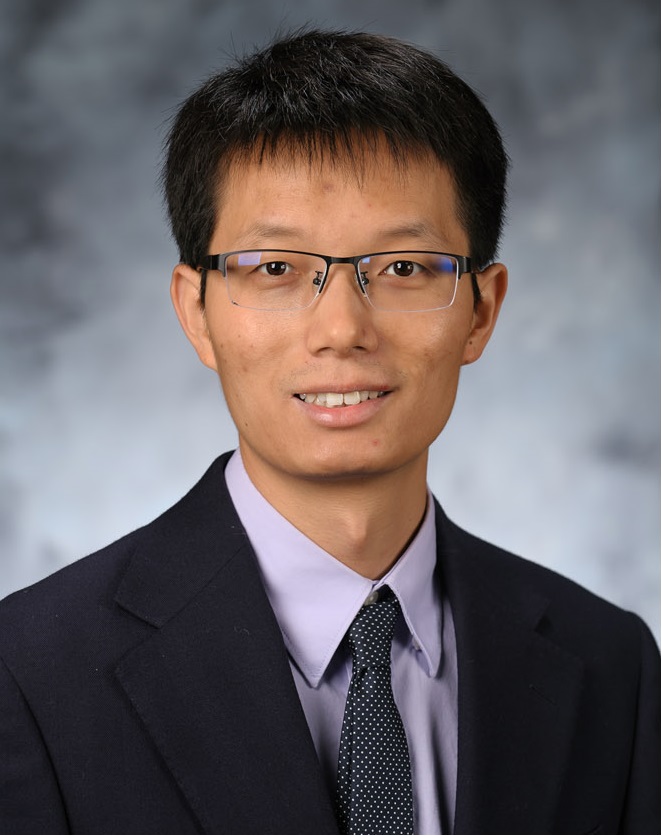}}]{Zheng Zhang}
Zheng Zhang [M'15] has been an Associate Professor of Electrical and Computer Engineering with the University of California at Santa Barbara (UCSB), since September 2023. He received his Ph.D in Electrical Engineering and Computer Science from the Massachusetts Institute of Technology (MIT), Cambridge, MA, in 2015, M.Phil from the University of Hong Kong in 2010, and B. Eng from Huazhong University of Science and Technology in 2008.
% His industrial experiences include Coventor Inc., Cambridge, MA, and Maxim-IC, Colorado Springs, CO, USA; academic visiting experiences include the University of California at San Diego, Brown University, and Politechnico di Milano, Milan, Italy; government laboratory experiences include the Argonne National Laboratory, Lemont, IL, USA. 
His research interests include uncertainty-aware design automation for electronics, photonics, and quantum circuits; small-data and data-free scientific machine learning for multi-physics design of 3D IC and chiplet, and tensor-compressed methods for sustainable training of large AI models and for resource-constraint on-device learning.
\end{IEEEbiography}


\vfill

\end{document}